\documentclass[11pt,a4paper]{article}

\usepackage[T1]{fontenc}
\usepackage[utf8]{inputenc}
\usepackage{CJKutf8}
\usepackage{times}

\usepackage[margin=2.4cm,headheight=38pt,headsep=18pt,footskip=28pt]{geometry}
\usepackage{setspace}
\usepackage[table,dvipsnames]{xcolor}
\usepackage{colortbl}
\usepackage{graphicx}
\usepackage{booktabs}
\usepackage{array}
\usepackage{multirow}
\usepackage{tabularx}

\usepackage{amsmath,amssymb}

\definecolor{AMapBlue}{HTML}{48B8F0}
\definecolor{AMapDeepBlue}{HTML}{0877C9}
\definecolor{AMapSky}{HTML}{EAF8FF}
\definecolor{AMapLine}{HTML}{9EDBFA}
\definecolor{TextBlack}{HTML}{20242A}

\usepackage[numbers,sort&compress]{natbib}

\usepackage{enumitem}
\setlist{nosep,leftmargin=1.5em}
\usepackage{caption}
\usepackage{fancyhdr}
\usepackage{lastpage}
\usepackage{titlesec}
\usepackage[most]{tcolorbox}
\usepackage{mdframed}
\usepackage[colorlinks=true,linkcolor=AMapDeepBlue,citecolor=AMapDeepBlue,urlcolor=AMapDeepBlue]{hyperref}

\newcommand{\reportname}{OneLA}
\newcommand{\reportlogo}{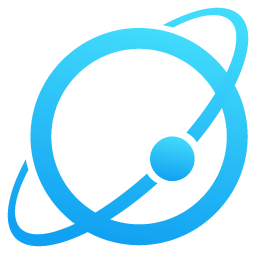}
\newcommand{\reportlogoheight}{18pt}

\newcommand{\orgname}{AMAP CV Lab \textbar{} Alibaba Group}
\newcommand{\reportcenter}[1]{\raisebox{\dimexpr(\depth-\height)/2\relax}{#1}}
\newlength{\reporttextbaselineoffset}
\newcommand{\reporttextalign}[1]{\raisebox{\reporttextbaselineoffset}{#1}}

\renewcommand{\headrulewidth}{1.2pt}
\renewcommand{\footrulewidth}{0.8pt}
\renewcommand{\headrule}{\hbox to\headwidth{\color{AMapBlue}\leaders\hrule height \headrulewidth\hfill}}
\renewcommand{\footrule}{\hbox to\headwidth{\color{AMapLine}\leaders\hrule height \footrulewidth\hfill}}

\titleformat{\section}
  {\Large\bfseries\color{AMapDeepBlue}}
  {\thesection}{0.75em}{}
  [\vspace{0.32em}{\color{AMapLine}\titlerule[0.6pt]}]
\titleformat{\subsection}
  {\large\bfseries\color{AMapDeepBlue}}
  {\thesubsection}{0.65em}{}
\titleformat{\subsubsection}
  {\normalsize\bfseries\color{AMapDeepBlue}}
  {\thesubsubsection}{0.55em}{}
\titlespacing*{\section}{0pt}{1.55em plus 0.25em minus 0.1em}{0.95em}
\titlespacing*{\subsection}{0pt}{1.25em plus 0.2em minus 0.1em}{0.55em}
\titlespacing*{\subsubsection}{0pt}{1.05em plus 0.15em minus 0.08em}{0.45em}

\newtcolorbox{abstractbox}{
  enhanced,
  breakable,
  colback=AMapSky,
  colframe=AMapBlue,
  coltitle=white,
  colbacktitle=AMapBlue,
  title=Abstract,
  fonttitle=\Large\bfseries,
  halign title=center,
  toptitle=5pt,
  bottomtitle=5pt,
  fontupper=\large,
  boxrule=0.8pt,
  arc=1.5mm,
  left=12pt,
  right=12pt,
  top=10pt,
  bottom=10pt,
  before skip=1.2em,
  after skip=0.5em
}

\newmdenv[
  backgroundcolor=gray!8,
  linewidth=0pt,
  innerleftmargin=10pt,
  innerrightmargin=10pt,
  innertopmargin=5pt,
  innerbottommargin=5pt,
  skipabove=5pt,
  skipbelow=5pt
]{cnblock}

\definecolor{DimAction}{HTML}{0877C9}
\definecolor{DimVisual}{HTML}{2E9E5B}
\definecolor{DimPhysics}{HTML}{E07B00}
\definecolor{DimMemory}{HTML}{8E44AD}

\usepackage{algorithm}
\usepackage{algpseudocode}

\usepackage{amsthm}

\usepackage{needspace}
\let\reportsection\section
\renewcommand{\section}{\Needspace{7\baselineskip}\reportsection}
\usepackage{afterpage}
\DeclareRobustCommand{\onela}{\textnormal{\bfseries\itshape OneLA}}
\newcommand{\R}{\mathbb{R}}
\newcommand{\E}{\mathbb{E}}
\hypersetup{pdftitle={OneLA: Learning Many Worlds from Latent Actions},pdfauthor={Ziyu Zhu, Zhaoxu Sun, Fan Jiang, Chiyu Wang, Junfeng Ni, Mu Xu}}

\begin{document}
\thispagestyle{fancy}
\begin{center}
{\bfseries\color{AMapDeepBlue}\LARGE OneLA: Learning Many Worlds \\ from Latent Actions\par}
\vspace{0.2em}
{\large\color{black}
Ziyu Zhu\textsuperscript{1}\quad Zhaoxu Sun\textsuperscript{1}\quad Fan Jiang\textsuperscript{\ensuremath{1,\ddagger}}\quad Chiyu Wang\textsuperscript{1}\\
Junfeng Ni\textsuperscript{\ensuremath{1,2,\S}}\quad Mu Xu\textsuperscript{1}\par}
\vspace{-0.1em}
{\small\color{black}\textsuperscript{1}AMAP, Alibaba Group\qquad\textsuperscript{2}Tsinghua University}
\end{center}
\vspace{-1em}
\begingroup
\renewcommand{\thefootnote}{}
\makeatletter
\long\def\@makefntext#1{\noindent#1}
\makeatother
\footnotetext[0]{\textsuperscript{\ensuremath{\ddagger}}Corresponding author.\\
\textsuperscript{\S}Work done during internship at AMAP, Alibaba Group.}
\endgroup
\begin{abstractbox}

Learning world models across diverse domains calls for a unified control space, yet action interfaces vary across domains and action annotations are often scarce. Latent actions learned directly from videos offer a promising way to establish such a space without explicit action supervision. However, learning shared latent actions across domains remains challenging, as variations in appearance, embodiment, and dynamics can entangle domain-specific visual cues with transferable transition structure.
To this end, we introduce \onela{}, a simple framework that learns continuous latent actions through a shared orthogonal dictionary and a common predictive objective, without task-specific designs or cross-domain alignment losses during latent action pretraining. These latent actions provide a shared conditioning interface for world models, with lightweight adapters mapping native control commands into the same space. Conditioned on latent actions and the current visual state, the world model predicts context-dependent transitions for controllable rollout. \onela{} improves action readout across domains and reduces action cycle-transfer rotation error by 24.8\% over the strongest evaluated baseline. A joint-training comparison further shows a 20.5\% improvement in Human Preference Score v3 (HPSv3) for real-scene generation.
These results suggest a path toward general-purpose world models that learn from diverse videos and support flexible control through latent action space.

The project page is available at \url{https://fantasy-amap.github.io/OneLA}.

\end{abstractbox}
\suppressfloats[t]

\section{Introduction}
\label{sec:introduction}

World models enable agents to anticipate the consequences of their actions and learn behavior through simulated interaction~\citep{ha2018worldmodels,hafner2025dreamer,alonso2024diamond}. Extending this capability across games, navigation, driving, and robotics calls for a shared action space that connects heterogeneous control interfaces to a common predictive model. Yet much of the available video data captures interactions without recording the underlying commands~\citep{gao2025adaworld,ye2025lapa}. A shared space learned from visual transitions could make these unlabeled videos useful for controllable world modeling.

Recent studies infer latent actions from videos to learn controllable world models~\citep{bruce2024genie,gao2025adaworld}. AdaWorld adapts world models to new control interfaces, while \citet{garrido2026wild} study action transfer across natural videos. Other work guides learning with agent localization~\citep{fechner2026masklam}, task descriptions or action-category priors~\citep{nikulin2026taskcentric,dai2026conla}, synchronized views~\citep{lee2026mvplam}, or labeled robot actions~\citep{jeong2026vila,alvarez2026actionsimilarity,liang2026clam,chen2026villax}. These signals help identify relevant visual changes, but are not uniformly available across sources. We study how a common regularizer can structure continuous actions across heterogeneous videos for a shared world-model interface.

\begin{figure}[!t]
    \centering
    \includegraphics[width=\linewidth]{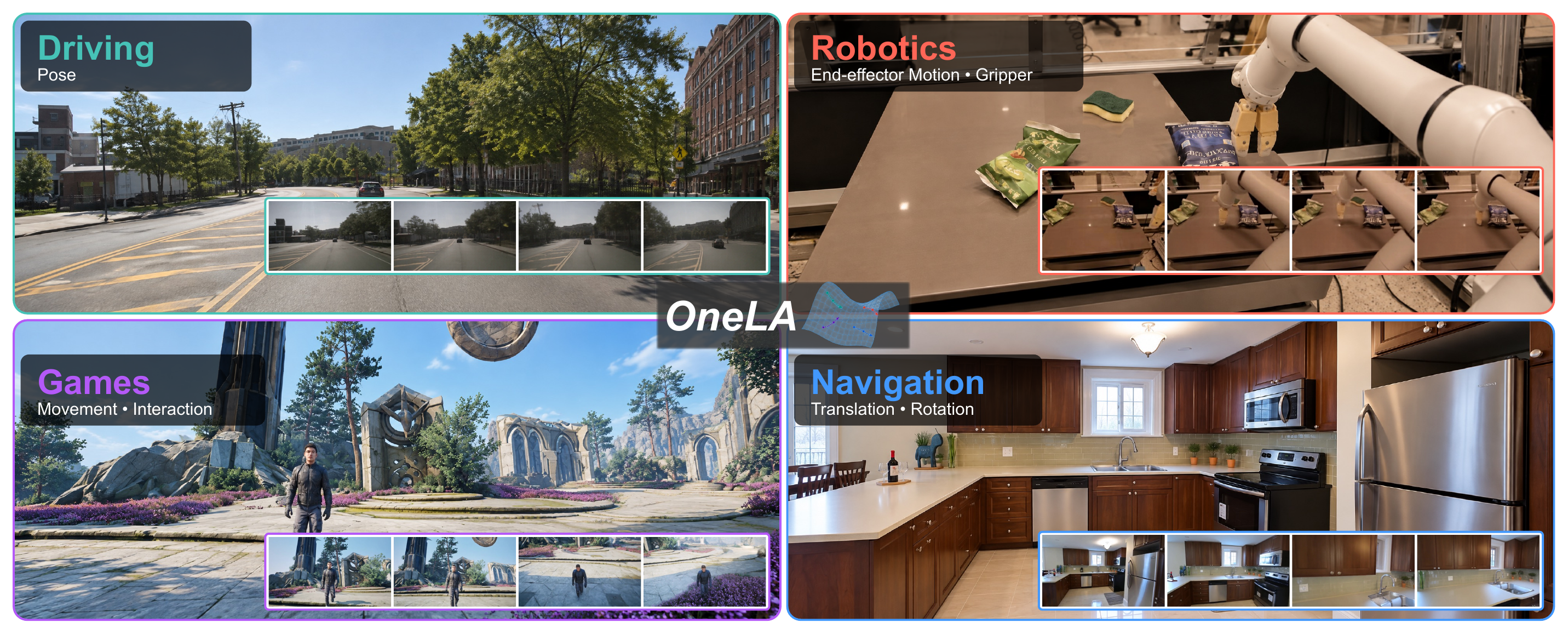}
    \caption{\textbf{\onela{} across many worlds.} A shared latent action interface connects video prediction to the different control signals of driving, robotics, games, and navigation. The inset boxes at the bottom of each panel show frames from model-generated rollouts.}
    \label{fig:onela-overview}
\end{figure}

Learning this shared interface presents two challenges. First, regularization must encourage \textit{reusable motion structure} without discarding information needed to represent diverse transitions. Frame prediction alone can retain scene-specific or uncontrolled changes alongside motion~\citep{zhang2025actually,nikulin2025laom}. Second, the interface must accommodate \textit{observed motion and control inputs}: video-derived latents describe realized transitions, whereas command-derived latents specify controls whose effects depend on the scene. A forward command, for example, may produce no motion when an obstacle blocks the agent. The world model must therefore use visual context to predict the outcome of either input.

We introduce \onela{}, a simple framework for learning a shared action interface from videos (Figure~\ref{fig:onela-overview}). A latent action model (LAM), formulated as an inverse dynamics model~\citep{garrido2026wild}, encodes frame pairs into continuous coefficients. We pretrain it across games, navigation, driving, and robotics without action labels, combining next-frame reconstruction with an orthonormal motion dictionary~\citep{wang2022lia}, Kullback--Leibler (KL) regularization, and sparsity. We then freeze the LAM and train lightweight multilayer perceptrons (MLPs) on labeled trajectories to align native controls to its latent space. Finally, we train a shared diffusion transformer conditioned on the initial frame and either video-derived pseudo-actions or mapped native controls. It jointly predicts video and realized-transition latent actions, using frame-pair codes from the frozen LAM as action targets.

We evaluate \onela{} on six datasets spanning four domains. It improves linear-probe macro-F1 over the strongest listed baseline by 5.30 points on Game and 5.63 on AgiBot (Table~\ref{tab:probes}). In world-model evaluation, it reduces action-cycle translation and rotation errors by 15.5\% and 24.8\% relative to OLaF and achieves the highest nuScenes peak signal-to-noise ratio (PSNR) and structural similarity (SSIM) among the evaluated methods (Table~\ref{tab:wm}).

Our main contributions are summarized as follows:
\begin{list}{\textbullet}{\setlength{\leftmargin}{2em}\setlength{\topsep}{0pt}\setlength{\itemsep}{0pt}\setlength{\parsep}{0pt}}
    \item We propose a shared latent action interface learned from unlabeled videos through an orthonormal motion dictionary and coefficient regularization.
    \item We introduce unified world-model training on labeled and unlabeled videos, combining native-control alignment with joint video and latent action prediction.
    \item We evaluate \onela{} across diverse domains, demonstrating improved action readout, cross-view transfer, and action-cycle consistency, with higher visual quality.
\end{list}

\afterpage{%
\begin{table}[!t]
\centering
\caption{\textbf{\onela{} learns latent actions across diverse domains with a simple shared objective.}}
\label{tab:lam-comparison}
\small
\setlength{\tabcolsep}{3pt}
\renewcommand{\arraystretch}{1.12}
\begin{tabular*}{\linewidth}{@{\extracolsep{\fill}} l c c c c c c @{}}
\toprule
\multirow{2}{*}[-2.5pt]{\textbf{Method}} & \multirow{2}{*}[-2.5pt]{\textbf{Domains}} & \multicolumn{5}{c}{\textbf{Auxiliary LAM objectives}} \\
\cmidrule(lr){3-7}
 & & Feat. & Act. & Temp. & Flow & Reg. \\
\midrule
AdaWorld~\citep{gao2025adaworld} & G, D, R & $\times$ & $\times$ & $\times$ & $\times$ & KL \\
VILA~\citep{jeong2026vila} & R & $\times$ & $\checkmark$ & $\times$ & $\times$ & Contr., struct. \\
LAOF~\citep{bu2026laof} & G, R & $\times$ & $\times$ & $\times$ & $\checkmark$ & --- \\
AC-LAM~\citep{wei2026aclam} & R & $\times$ & $\checkmark$ & $\checkmark$ & $\times$ & VQ \\
Olaf-World~\citep{jiang2026olaf} & G, N & $\checkmark$ & $\times$ & $\times$ & $\times$ & KL \\
DiLA~\citep{zhang2026dila} & G, N, R & $\times$ & $\times$ & $\checkmark$ & $\times$ & Norm, var. \\
\midrule
\onela{} (ours) & \textbf{G, N, D, R} & $\times$ & $\times$ & $\times$ & $\times$ & $\mathrm{KL}+\ell_1$ \\
\bottomrule
\end{tabular*}
\par\smallskip
\begin{minipage}{\linewidth}
\footnotesize
G: games; N: navigation; D: driving; R: robot manipulation. Feat.: feature/effect alignment; Act.: action/proprioceptive supervision; Temp.: temporal consistency, composition, or inversion; Flow: optical-flow prediction; Reg.: latent regularization; Contr., struct.: contrastive and structural regularization. VQ: vector quantization; Norm, var.: norm and variance regularization.
\end{minipage}
\end{table}
}
\section{Related Work}
\label{sec:related}

\subsection{Latent Actions from Video}
Recent studies learn latent actions from visual transitions for controllable generation~\citep{bruce2024genie,gao2025adaworld} and policy pretraining~\citep{ye2025lapa,bu2025univla,chen2025moto,tharwat2025lawm}. A central difficulty is that predictive codes can retain appearance and uncontrolled changes alongside motion~\citep{zhang2025actually,nikulin2025laom}. Information bottlenecks restrict what the code can retain, trading off action abstraction against predictive detail. Discrete tokenizers~\citep{bruce2024genie,ye2025lapa} and regularized continuous spaces~\citep{gao2025adaworld,garrido2026wild} make different choices in this trade-off, whose effects on action abstraction and prediction have motivated systematic study~\citep{yoshimoto2026flexlam,bu2026matters}. Beyond capacity, learning transferable actions requires deciding which visual changes should have similar representations. Methods guide this choice through agent masks, optical-flow targets, or action supervision~\citep{fechner2026masklam,bu2026laof,jeong2026vila}; others impose content--structure separation~\citep{zhang2026dila}, compositional or cycle consistency~\citep{wei2026aclam,chen2026cyclemimic}, or alignment to video-feature changes~\citep{jiang2026olaf}. To structure continuous actions across sources, \onela{} uses an orthonormal dictionary~\citep{wang2022lia} that gives KL and sparsity regularization a consistent geometric scale. It learns through a common reconstruction objective without additional supervision or alignment targets during latent action pretraining. Table~\ref{tab:lam-comparison} compares auxiliary objectives beyond transition prediction.

\subsection{World Models}
World models support control through video simulation~\citep{valevski2025gamengen,yu2025gamefactory,gao2024vista,zhao2025drivedreamer2,russell2025gaia2} or feature-space prediction for planning~\citep{zhou2025dinowm,assran2025vjepa2}. Broad video pretraining supplies visual and dynamics priors~\citep{nvidia2025cosmos,huang2026vid2world}, while geometry and memory help maintain consistent rollouts~\citep{chen2025deepverse,sun2025worldplay,zhu2025aether}. Reusing these predictive models across domains also requires reconciling different control interfaces. Latent actions provide an intermediate representation: they connect unlabeled video pretraining to downstream commands~\citep{gao2025adaworld,gao2026dreamdojo} and make video-derived motion useful for imitation and feature-space dynamics~\citep{wang2026glam,zhang2026rla}. Recent unified models further accommodate diverse controls through shared and modality-specific experts~\citep{fang2026worldscape} or partially revealed visual trajectories~\citep{alzayer2026masked}. \onela{} couples a geometrically regularized action space with joint multi-domain video and action prediction, enabling native-command control and reference-action transfer through one shared world-model backbone.

\section{\onela{}}
\label{sec:method}

\subsection{Formulation}
\label{sec:formulation}
We represent a visual transition $(x_t,x_{t+\Delta})$ by a latent action $z_t\in\mathcal{Z}=\R^d$, where $\Delta$ is the frame interval. The LAM infers $z_t$ from the frame pair, while a decoder reconstructs $x_{t+\Delta}$ from $x_t$ and $z_t$.

\begin{figure}[!ht]
    \centering
    \includegraphics[width=\linewidth]{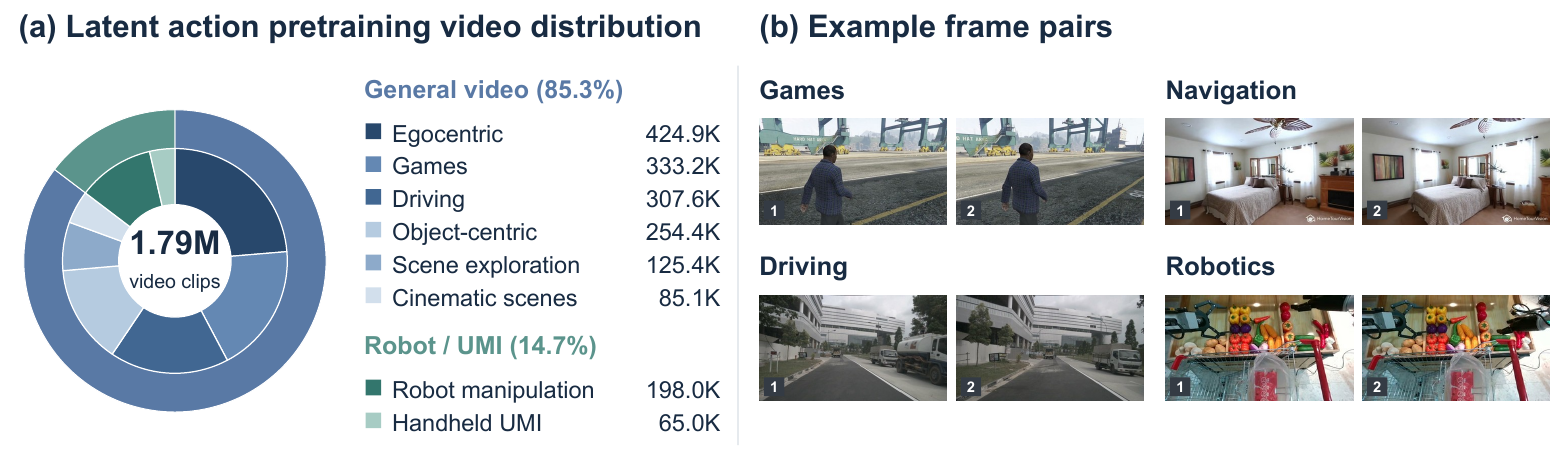}
    \caption{\textbf{Video distribution for latent action pretraining.} (a) Clip counts grouped by video category. (b) Example frame pairs from games, navigation, driving, and robotics.}
    \label{fig:onela-data}
\end{figure}

Training has three stages. We first learn the LAM with a shared orthonormal motion dictionary from videos without action labels (Section~\ref{sec:orthogonal}). We then freeze it and train lightweight MLPs on labeled trajectories to map native controls $u_t^{(k)}\in\mathcal{U}_k$ into $\mathcal{Z}$. Finally, we train a shared video world model,
\begin{equation}
 p_\theta(x_{1:T}\mid x_0,c_{1:L}),\qquad c_\ell\in\mathcal{Z},
 \label{eq:wm}
\end{equation}
conditioned on the initial frame $x_0$ and latent actions $c_{1:L}$. For world-model clips, we index actions by their destination frames: $\bar z_\ell=\mu_\phi(x_{\ell-1},x_\ell)$ and $L=T$; the adapter groups four actions per video-latent transition. These conditions are either LAM-inferred pseudo-actions or, for labeled clips, MLP-mapped native controls. The model jointly predicts video and realized-transition latent actions, using the frozen LAM to supply action targets (Section~\ref{sec:worldmodel}).

\subsection{Latent Action Pretraining Data}
\label{sec:method-data}
Our pretraining collection contains 1.79M video clips spanning games, navigation, driving, egocentric interactions, and robot manipulation. We sample temporally ordered frame pairs to learn from camera motion, agent movement, and object interaction without using action labels. A common LAM, motion dictionary, and reconstruction decoder process all sources.
Figure~\ref{fig:onela-data} groups the collection by video category and shows example frame pairs.

\subsection{Orthogonal Latent Action Learning}
\label{sec:orthogonal}
We express visual transitions as combinations of learned motion directions. The LAM predicts a compact coefficient vector $z_t$, which a shared dictionary embeds for next-frame reconstruction. QR factorization keeps dictionary directions orthonormal, giving coefficient regularization a fixed geometric scale. KL and sparsity regularization then control the information and combinations used to explain visual changes.

\paragraph{Inferring transition coefficients.}
The LAM encodes a frame pair with a spatiotemporal transformer~\citep{bruce2024genie} and a learned action token. It predicts a diagonal Gaussian posterior,
\begin{equation}
 q_\phi(z_t\mid x_t,x_{t+\Delta})
 =\mathcal{N}\!\left(\mu_\phi(x_t,x_{t+\Delta}),
          \operatorname{diag}(\sigma_\phi^2(x_t,x_{t+\Delta}))\right),
 \label{eq:inverse}
\end{equation}
where $z_t\in\R^d$ specifies the coefficients of a transition. During pretraining, we sample $z_t$ by reparameterization. After pretraining, we use the posterior mean $\bar z_t=\mu_\phi(x_t,x_{t+\Delta})$ as the deterministic latent action for command alignment and world modeling.

\paragraph{A shared orthonormal motion dictionary.}
To regularize coefficients on a consistent geometric scale, we use an orthonormal motion dictionary~\citep{wang2022lia} with $d$ directions in an embedding space of dimension $h\geq d$. In our implementation, we obtain them by reduced QR factorization of a trainable matrix $W\in\R^{h\times d}$, assuming full column rank:
\begin{equation}
 W=QR,\qquad Q=[q_1,\ldots,q_d],\qquad Q^\top Q=I_d.
 \label{eq:qr}
\end{equation}
The columns $q_i$ are unit-norm, mutually orthogonal directions. A transition embedding is their weighted sum, $a_t=Qz_t=\sum_{i=1}^{d}z_{t,i}q_i$. Thus, different transitions can combine the same directions with different coefficients. Gradients pass through the factorization to update $W$, so the dictionary is learned jointly with the LAM and reconstruction decoder.

The decoder receives both the current observation and this transition embedding. Specifically, we add $Qz_t$ to each projected current-frame patch $p_j(x_t)\in\R^h$ and reconstruct the next frame with a spatial transformer $F_\psi$:
\begin{equation}
 \widetilde p_{t,j}=p_j(x_t)+Qz_t,
 \qquad
 \widehat x_{t+\Delta}=F_\psi(\{\widetilde p_{t,j}\}_j).
 \label{eq:decoder}
\end{equation}
This combines a global transition condition with local visual content. The same dictionary and decoder process every data source. The resulting coefficients $z_t$ form the shared interface for world modeling; $Qz_t$ is the embedding used by the pretraining decoder.

\paragraph{Learning motion structure through regularization.}
QR provides a simple form of structural regularization at the embedding interface (Appendix~\ref{app:geometry}):
\begin{equation}
 \|Qz\|_2=\|z\|_2,\qquad \|Qz-Qz'\|_2=\|z-z'\|_2.
 \label{eq:isometry}
\end{equation}
Consequently, dictionary directions cannot become collinear or be arbitrarily enlarged to offset smaller coefficients. This removes a scale ambiguity between the coefficients and dictionary columns, giving coefficient regularization a consistent geometric scale. It constrains how transitions are represented, without assigning semantic meanings to the learned directions. We jointly train the LAM, dictionary, and reconstruction decoder with
\begin{equation}
 \mathcal{L}_{\mathrm{LAM}}=
 \E_{(x_t,x_{t+\Delta}),\,z_t\sim q_\phi}
       \!\left[\|x_{t+\Delta}-\widehat x_{t+\Delta}\|_2^2\right]
 +\beta\,\E\!\left[D_{\mathrm{KL}}(q_\phi\|\mathcal{N}(0,I))\right]
 +\frac{\lambda}{d}\,\E\!\left[\|\mu_\phi\|_1\right].
 \label{eq:lamloss}
\end{equation}
Reconstruction loss is averaged over valid pixels. The KL term limits information in the coefficients, while the $\ell_1$ term encourages sparse posterior means; $\beta$ and $\lambda$ weight the two terms. All video sources share this objective without action labels or domain-specific prediction targets.

\subsection{World Model Training}
\label{sec:worldmodel}
\begin{figure}[t]
    \centering
    \includegraphics[width=\linewidth]{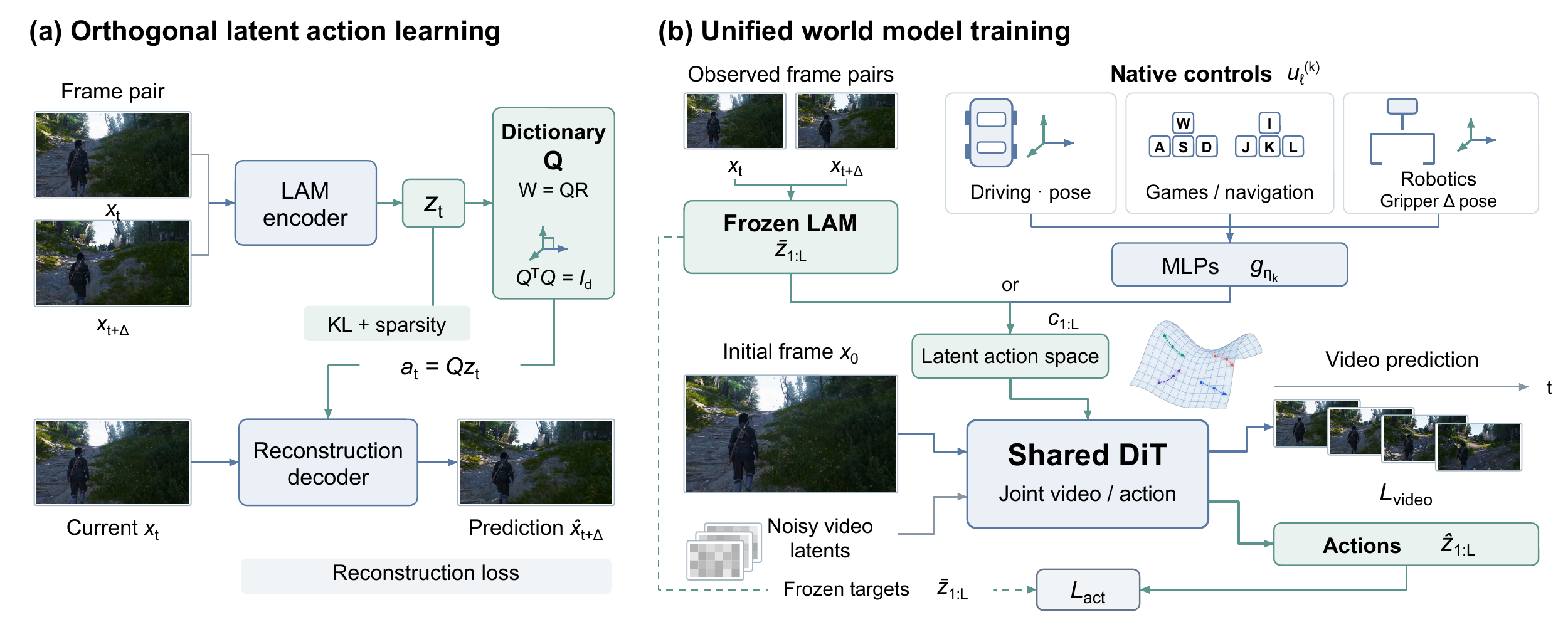}
    \caption{\textbf{Overview of \onela{}.} (a) Orthogonal latent action learning from frame pairs. (b) Joint video and action prediction conditioned on inferred latent actions or mapped native controls.}
    \label{fig:onela-method}
\end{figure}

We use the diffusion transformer (DiT)~\citep{peebles2023dit} from Wan2.2-TI2V-5B~\citep{wan2025,wan2025modelcard} for video generation conditioned on an initial frame and latent actions (Figure~\ref{fig:onela-method}).

For each labeled clip, we sample either frozen-LAM conditions $c_\ell=\bar z_\ell$ or MLP-mapped controls $c_\ell=g_{\eta_k}(u_\ell^{(k)})$. Unlabeled clips use only the former. A trainable adapter temporally aligns and projects these coefficients into the DiT. The pretrained command MLPs are then optimized jointly with the video model and adapter, while the LAM remains frozen.

We jointly predict video and latent actions, a strategy also explored in robot learning~\citep{cheang2024gr2,routray2026vipra}. For both conditioning routes, the frozen LAM extracts action targets $\bar z_{1:L}$ from observed frame pairs to supervise realized transitions. The objective is
\begin{equation}
 \mathcal{L}_{\mathrm{WM}}
 =\mathcal{L}_{\mathrm{video}}+\lambda_{\mathrm{act}}\mathcal{L}_{\mathrm{act}},
 \label{eq:joint-loss}
\end{equation}
where $\mathcal{L}_{\mathrm{video}}$ is the video flow-matching loss~\citep{lipman2023flow}, and $\mathcal{L}_{\mathrm{act}}$ combines SmoothL1 regression and cosine distance between predicted and target coefficients.

\section{Experiments}
\label{sec:experiments}

\subsection{Experimental Setting}
\label{sec:setup}
\paragraph{Datasets.}
Latent action pretraining uses the collection in Section~\ref{sec:method-data}. World-model training uses 11 sources, including ABot-World~\citep{jiang2026abotworld0}, nuScenes~\citep{caesar2020nuscenes}, Sekai~\citep{li2025sekai}, RealEstate10K~\citep{zhou2018stereo}, and RT-1~\citep{brohan2023rt1}. Representation tests cover four domains, including AgiBot~\citep{agibot2025}; Game is a held-out ABot-World split. We evaluate video prediction on MIND~\citep{ye2026mind}, nuScenes, and RT-1, and action cycle transfer between game and real scenes. For \onela{}, MIND, nuScenes, and RT-1 prediction uses the initial frame and MLP-mapped native commands, without target-frame latent conditions; cycle transfer instead extracts latent actions from the source video.

\paragraph{Implementation.}
The LAM latent dimension is 32. Its default pretraining schedule is configured for 50k steps, a global batch of 640 frame pairs, and an initial learning rate of $5\times10^{-5}$. Across all quantitative benchmarks in Table~\ref{tab:wm}, \onela{} uses the same trained Wan2.2-TI2V-5B DiT weights~\citep{wan2025modelcard,jiang2026abotworld0}, frozen LAM, and latent action space; only the command MLPs differ by control interface. We evaluate after 6k fine-tuning updates with a global batch of 16 clips and an initial learning rate of $10^{-5}$. Both stages use AdamW on 16 GPUs with cosine learning-rate decay. Joint training uses $\lambda_{\mathrm{act}}=0.05$. Further details are in the appendix.

\paragraph{Baselines.}
Using released LAM weights, we compare frozen representations with OLaF~\citep{jiang2026olaf}, its variational autoencoder (VAE) without alignment, AdaWorld~\citep{gao2025adaworld}, DiLA~\citep{zhang2026dila}, and RLA-WM~\citep{zhang2026rla}; for OLaF, AdaWorld, and DiLA, we fine-tune a world model for each released frozen LAM using the same Wan backbone, data mixture, and optimization settings as \onela{} in the Implementation paragraph above. Within each benchmark, methods use the same evaluation cases, readouts, and metrics, retaining native dimensions and method-specific preprocessing. Additional baseline details are provided in the appendix.

\subsection{Main Results}
\label{sec:main_results}
\paragraph{Action representation.}
\label{sec:probes}
\label{sec:transfer}
Table~\ref{tab:probes} shows that \onela{} achieves the best F1 on most datasets and the second-best on the rest under both full-representation linear probing and single-dimension readout. The latter uses one scalar latent dimension per action label. On Game, the single-dimension readout achieves 65.26 F1 versus 67.08 using all dimensions, indicating that action information is concentrated in individual coordinates. MIND cross-view F1 reaches 78.31 and 72.88, exceeding the best baselines by 10.66 and 5.53 points without target-view labels for probe fitting.

\begin{table}[!b]
\vspace{-10pt}
\centering
\caption{\textbf{Action information in frozen LAM representations.} Macro-F1 (\%, $\uparrow$) for the three readout settings. Boldface and underlining indicate the best and second-best scores.}
\label{tab:probes}
\small
\renewcommand{\arraystretch}{0.95}
\setlength{\aboverulesep}{1.4pt}
\setlength{\belowrulesep}{1.8pt}
\setlength{\tabcolsep}{3pt}
\begin{tabular*}{\linewidth}{@{\extracolsep{\fill}}lcccccc@{}}
\toprule
\multicolumn{7}{l}{\textbf{(a) Linear probe (all dimensions)}} \\
\midrule
Method & Game & Sekai & MIND & nuScenes & RT-1 & AgiBot \\
\midrule
OLaF~\citep{jiang2026olaf} & \underline{61.78} & \textbf{53.84} & \underline{81.41} & \underline{87.22} & 60.02 & \underline{48.90} \\
AdaWorld~\citep{gao2025adaworld} & 58.95 & 50.63 & 79.68 & 80.84 & \textbf{61.58} & 45.41 \\
OLaF w/o align (VAE)~\citep{jiang2026olaf} & 56.43 & 48.02 & 75.57 & 78.39 & 58.59 & 43.87 \\
DiLA~\citep{zhang2026dila} & 40.55 & 34.90 & 57.94 & 59.05 & 56.32 & 46.53 \\
RLA-WM~\citep{zhang2026rla} & 45.26 & 26.33 & 64.48 & 49.58 & 52.20 & 39.09 \\
\midrule
\onela{} (ours) & \textbf{67.08} & \underline{53.60} & \textbf{89.21} & \textbf{87.53} & \underline{60.85} & \textbf{54.53} \\
\bottomrule
\end{tabular*}
\par\smallskip

\begin{tabular*}{\linewidth}{@{\extracolsep{\fill}}lccccccc@{}}
\toprule
 & \multicolumn{5}{c}{\textbf{(b) Single-dimension readout}} & \multicolumn{2}{c}{\textbf{(c) Cross-view transfer}} \\
\cmidrule(lr){2-6}\cmidrule(l){7-8}
Method & Game & Sekai & MIND & RT-1 & AgiBot & $1\mathrm{st}{\to}3\mathrm{rd}$ & $3\mathrm{rd}{\to}1\mathrm{st}$ \\
\midrule
OLaF~\citep{jiang2026olaf} & \underline{55.09} & 47.40 & \underline{70.57} & 57.82 & 36.30 & 64.72 & 63.92 \\
AdaWorld~\citep{gao2025adaworld} & 53.98 & \underline{49.31} & 68.55 & \textbf{59.75} & 38.45 & 60.84 & \underline{67.35} \\
OLaF w/o align (VAE)~\citep{jiang2026olaf} & 53.49 & 41.89 & 58.08 & 57.12 & 35.92 & \underline{67.65} & 62.32 \\
DiLA~\citep{zhang2026dila} & 33.36 & 30.81 & 34.58 & 56.10 & 38.93 & 37.67 & 37.31 \\
RLA-WM~\citep{zhang2026rla} & 29.72 & 28.07 & 27.29 & 56.90 & \underline{39.65} & 46.42 & 46.13 \\
\midrule
\onela{} (ours) & \textbf{65.26} & \textbf{56.91} & \textbf{77.98} & \underline{59.48} & \textbf{45.90} & \textbf{78.31} & \textbf{72.88} \\
\bottomrule
\end{tabular*}
\end{table}

\paragraph{Video prediction.}
\label{sec:wm_eval}
\onela{} reduces MIND translation/rotation RPE by 15.2\%/7.6\% over the best baselines and leads nuScenes PSNR/SSIM (Table~\ref{tab:wm}). On RT-1, \onela{} ranks second across PSNR, SSIM, and LPIPS. Across 40 action-transfer cycles, which transfer source motion to a new scene and back, \onela{} reduces translation and rotation RPE by 15.5\% and 24.8\% relative to OLaF, the strongest comparator, indicating better round-trip preservation of source motion.

\begin{table}[t]
\centering
\caption{\textbf{Video prediction and action cycle transfer.} Video prediction uses mapped native commands; cycle transfer uses video-derived latent actions. SSIM~\citep{wang2004ssim} and LPIPS~\citep{zhang2018lpips} measure fidelity; IQ is MUSIQ-SPAQ~\citep{ke2021musiq}; RPE denotes translation ($t$) and rotation ($r$, degrees) error. Boldface and underlining indicate the best and second-best scores.}
\label{tab:wm}
\small
\setlength{\tabcolsep}{3pt}
\begin{tabular*}{\linewidth}{@{\extracolsep{\fill}}lcccccc@{}}
\toprule
 & \multicolumn{3}{c}{\textbf{nuScenes}} & \multicolumn{3}{c}{\textbf{RT-1}} \\
\cmidrule(lr){2-4}\cmidrule(l){5-7}
Method & PSNR $\uparrow$ & SSIM $\uparrow$ & LPIPS $\downarrow$ & PSNR $\uparrow$ & SSIM $\uparrow$ & LPIPS $\downarrow$ \\
\midrule
OLaF~\citep{jiang2026olaf} & \underline{15.92} & 0.601 & \textbf{0.525} & \textbf{18.78} & \textbf{0.785} & \textbf{0.321} \\
AdaWorld~\citep{gao2025adaworld} & 10.08 & 0.208 & 0.801 & 17.16 & 0.743 & 0.367 \\
DiLA~\citep{zhang2026dila} & 15.49 & \underline{0.604} & 0.538 & 17.28 & 0.734 & 0.370 \\
\midrule
\onela{} & \textbf{16.13} & \textbf{0.625} & \underline{0.533} & \underline{17.78} & \underline{0.771} & \underline{0.338} \\
\bottomrule
\end{tabular*}
\par\smallskip

\begin{tabular*}{\linewidth}{@{\extracolsep{\fill}}lccccc@{}}
\toprule
 & \multicolumn{3}{c}{\textbf{MIND}} & \multicolumn{2}{c}{\textbf{Action cycle transfer}} \\
\cmidrule(lr){2-4}\cmidrule(l){5-6}
Method & IQ ($\times100$) $\uparrow$ & RPE$_t$ $\downarrow$ & RPE$_r$ $\downarrow$ & RPE$_t$ $\downarrow$ & RPE$_r$ $\downarrow$ \\
\midrule
OLaF~\citep{jiang2026olaf} & \underline{62.11} & 0.045 & \underline{0.598} & \underline{0.092} & \underline{0.496} \\
AdaWorld~\citep{gao2025adaworld} & \textbf{62.30} & \underline{0.044} & 0.601 & 0.102 & 0.535 \\
DiLA~\citep{zhang2026dila} & 60.15 & 0.051 & 0.945 & 0.099 & 0.666 \\
\midrule
\onela{} & 61.19 & \textbf{0.037} & \textbf{0.553} & \textbf{0.077} & \textbf{0.373} \\
\bottomrule
\end{tabular*}
\end{table}

\subsection{Ablation Study}
\label{sec:ablations}
\paragraph{How are cross-domain action prototypes organized?}
Figure~\ref{fig:qr-cross-domain} (enlarged in Appendix~\ref{app:enlarged-qr}) compares checkpoints with and without QR at sparsity weight $0.01$. With action-balanced samples, the gap between same- and different-action cosine similarities increases from $0.715$ to $0.836$ across MIND views and from $0.140$ to $0.181$ across robots, indicating greater separation between matched and unmatched action-label prototypes. With QR, prototypes sharing the same action label have higher average similarity across domains, supporting better alignment of action-label-conditioned latent representations.

\begin{figure}[!htbp]
\centering
\includegraphics[width=\linewidth]{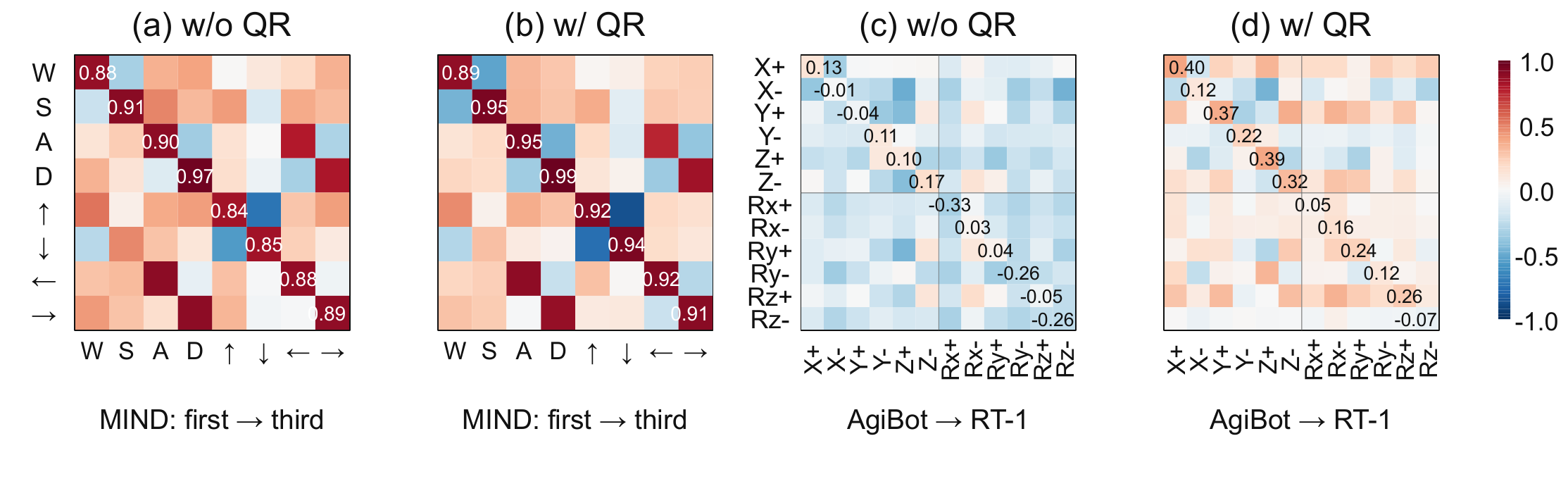}
\caption{\textbf{Cross-domain action prototypes with and without QR.} Cosine similarity between action-conditioned posterior means, across MIND viewpoints (left) and AgiBot/RT-1 robots (right). Rows and columns denote action classes.}
\label{fig:qr-cross-domain}
\end{figure}

\begin{figure}[!htbp]
\centering
\includegraphics[width=\linewidth]{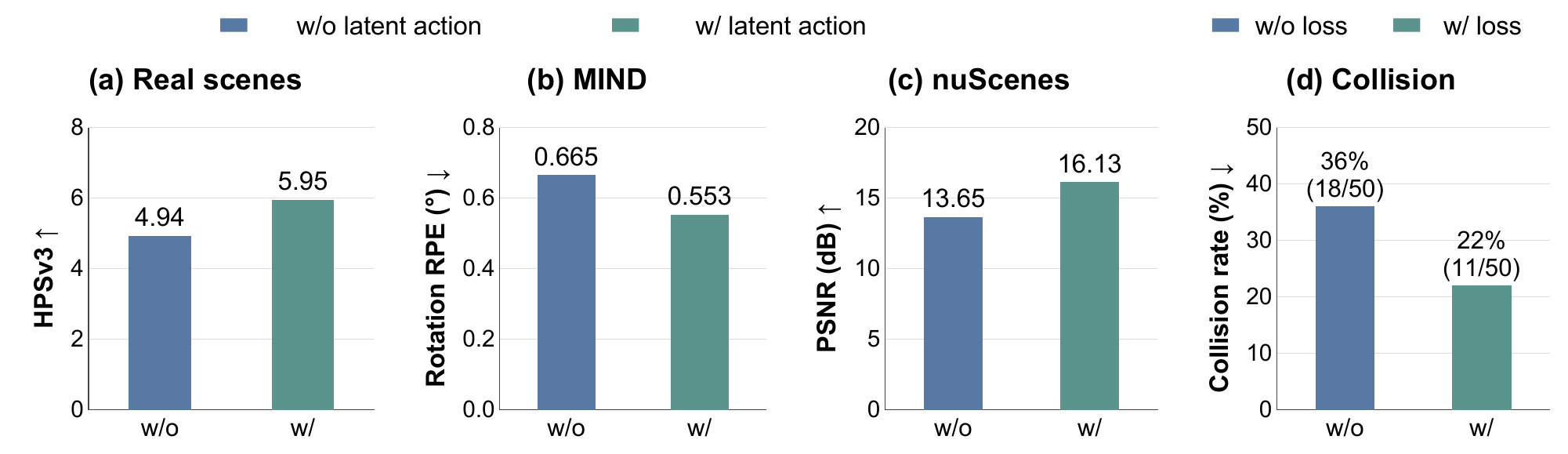}
\caption{\textbf{World-model training comparisons.} (a--c) With identical data and training procedures, w/ and w/o latent action use LAM-inferred and zero action vectors for unlabeled videos, respectively. Real-scene HPSv3 averages 16 generated frames per video over 30 scenes with fixed per-case target-scene text captions. (d) Latent-action prediction loss, with 50 cases per setting.}
\label{fig:wm-training-ablation}
\end{figure}

\paragraph{How do latent actions affect world-model training?}
Figure~\ref{fig:wm-training-ablation}(a--c) compares world-model training with and without latent actions for unlabeled videos. Both settings train the command MLPs under the same data and optimization setup, differing only in whether unlabeled videos receive LAM-inferred latent actions or zero vectors. Using these conditions improves real-scene HPSv3~\citep{ma2025hpsv3} from 4.94 to 5.95, MIND rotation RPE from 0.665 to 0.553, and nuScenes PSNR from 13.65 to 16.13. These results show that latent action conditioning can improve visual quality and motion prediction. Separately, panel~(d) shows that adding the latent-action prediction loss reduces human-verified collisions from 36\% (18/50) to 22\% (11/50).

\begin{samepage}
\paragraph{Does scaling improve action readout?}
Increasing each encoder and decoder stack from 6 to 12 layers at the same update budget improves Game, nuScenes, and RT-1 F1 by 6.44, 10.30, and 1.35 points (Table~\ref{tab:model-data-scaling}(a)). The 24-layer model scores highest among the reported models. Data scaling improves all three datasets, with Game peaking at $2\times$ and nuScenes/RT-1 at $3\times$ (Table~\ref{tab:model-data-scaling}(b)).
\par
\end{samepage}

\begin{table}[!htbp]
\centering
\caption{\textbf{Model size and data amount.} Linear-probe macro-F1 (\%, $\uparrow$). Data amount denotes cumulative sampled frame pairs. Boldface and underlining mark best and second-best scores.}
\label{tab:model-data-scaling}
\small
\begin{minipage}[t]{0.54\linewidth}
\centering
\textbf{(a) Model size}\par\smallskip
\setlength{\tabcolsep}{3pt}
\begin{tabular*}{\linewidth}{@{\extracolsep{\fill}}rcccc@{}}
\toprule
Layers & Params (M) & Game & nuScenes & RT-1 \\
\midrule
6 & 180.9 & 55.94 & 71.29 & 57.81 \\
12 & 357.2 & \underline{62.38} & \underline{81.59} & \underline{59.16} \\
24 & 709.8 & \textbf{65.35} & \textbf{84.02} & \textbf{60.05} \\
\bottomrule
\end{tabular*}
\end{minipage}\hfill
\begin{minipage}[t]{0.43\linewidth}
\centering
\textbf{(b) Data amount}\par\smallskip
\setlength{\tabcolsep}{3pt}
\begin{tabular*}{\linewidth}{@{\extracolsep{\fill}}cccc@{}}
\toprule
Data & Game & nuScenes & RT-1 \\
\midrule
$1\times$ & 64.04 & 81.63 & 59.26 \\
$2\times$ & \textbf{67.60} & \underline{84.38} & \underline{60.26} \\
$3\times$ & \underline{66.06} & \textbf{87.64} & \textbf{60.71} \\
\bottomrule
\end{tabular*}
\end{minipage}
\end{table}

\paragraph{Does a broader data mixture help?}
Expanding game-only training to the full mixture improves F1 on all six datasets, including Game (58.41 to 66.34) and AgiBot (39.18 to 54.42; Table~\ref{tab:ablation-data-types}). Sekai drops from 55.60 to 53.87 in the robot-inclusive setting, consistent with a possible trade-off in domain-specific readout.

\begin{table}[!htbp]
\centering
\caption{\textbf{Training-data composition.} Linear-probe macro-F1 (\%, $\uparrow$). Boldface and underlining indicate the best and second-best scores.}
\label{tab:ablation-data-types}
\small
\setlength{\tabcolsep}{3pt}
\begin{tabular*}{\linewidth}{@{\extracolsep{\fill}}cccc@{\hspace{16pt}}cccccc@{}}
\toprule
\multicolumn{4}{c}{Training sources} & \multicolumn{6}{c}{Evaluation datasets} \\
\cmidrule(lr){1-4}\cmidrule(lr){5-10}
\shortstack{G\\Games} & \shortstack{V\\Real videos} & \shortstack{D\\Driving} & \shortstack{R\\Robotics} & Game & Sekai & MIND & nuScenes & RT-1 & AgiBot \\
\midrule
$\checkmark$ & $\times$ & $\times$ & $\times$ & 58.41 & 44.60 & 71.16 & 73.64 & 58.79 & 39.18 \\
$\checkmark$ & $\checkmark$ & $\times$ & $\times$ & 64.82 & \underline{54.44} & 82.49 & 83.84 & 58.76 & 49.79 \\
$\checkmark$ & $\checkmark$ & $\checkmark$ & $\times$ & \underline{65.14} & \textbf{55.60} & \underline{83.74} & \underline{84.93} & \underline{58.89} & \underline{50.79} \\
$\checkmark$ & $\checkmark$ & $\checkmark$ & $\checkmark$ & \textbf{66.34} & 53.87 & \textbf{84.58} & \textbf{87.49} & \textbf{61.08} & \textbf{54.42} \\
\bottomrule
\end{tabular*}
\end{table}

\subsection{Qualitative Results}
\label{sec:qualitative}
\begin{figure}[!t]
\centering
\includegraphics[width=0.90\linewidth]{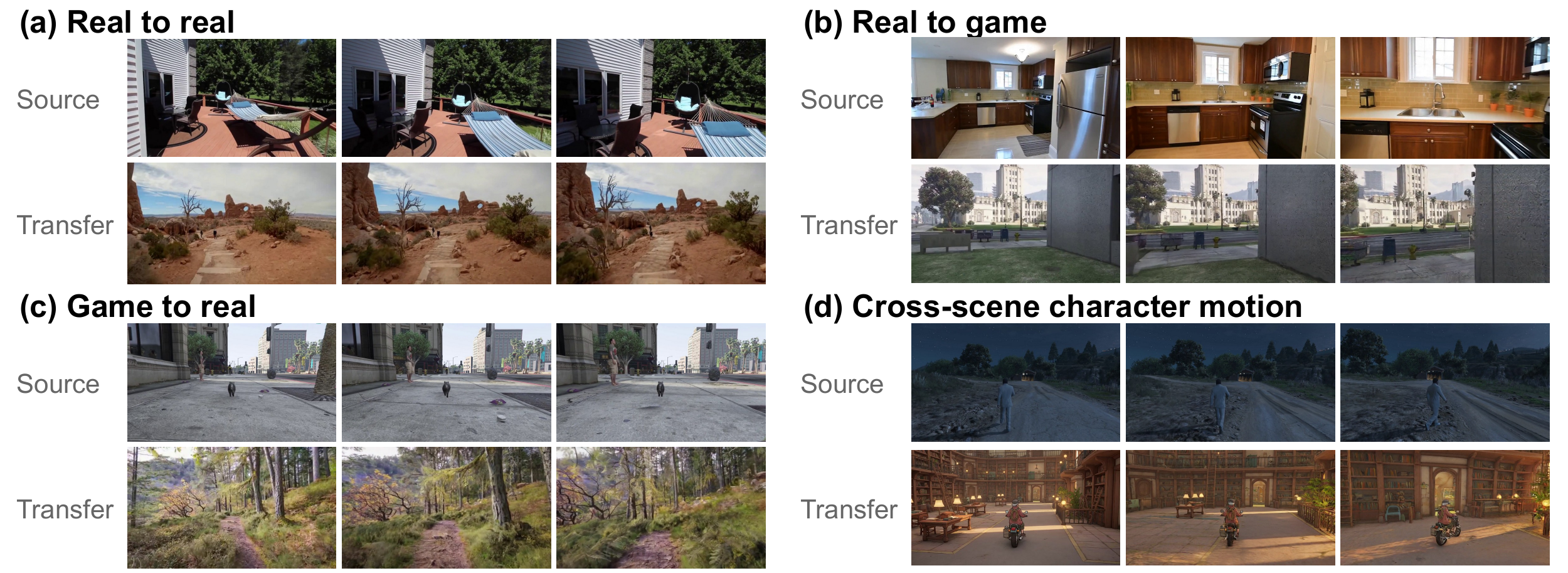}
\caption{\textbf{Reference-based action transfer.} Source (top) and Transfer (bottom) show matched relative frame indices; each transfer rollout starts from its first displayed frame.}
\label{fig:qualitative-action-transfer}
\end{figure}

\begin{figure}[!t]
\centering
\includegraphics[width=0.90\linewidth]{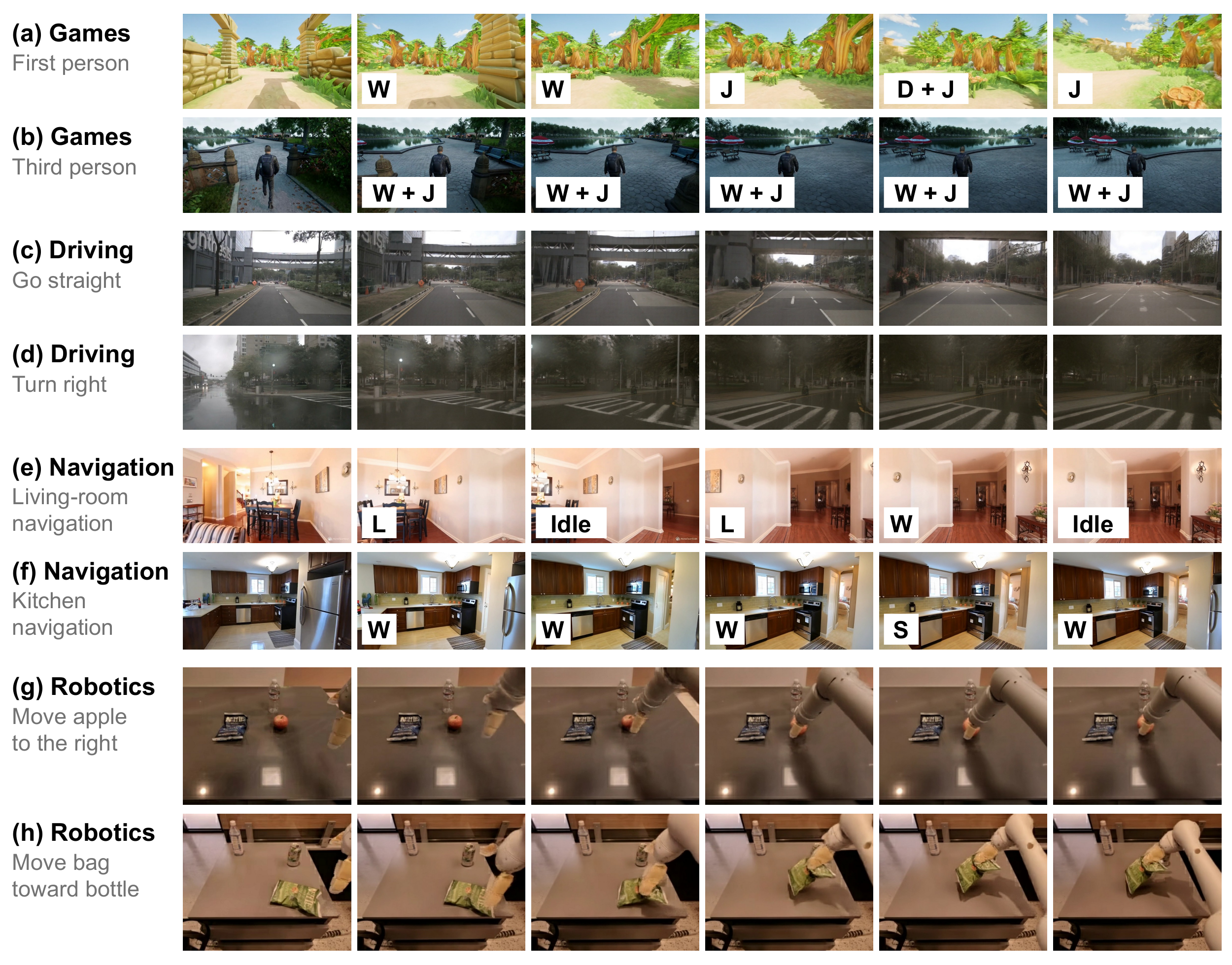}
\caption{\textbf{Qualitative rollouts across four domains.} Each row shows an input frame followed by five generated frames.}
\label{fig:qualitative-rollouts}
\end{figure}

Figures~\ref{fig:qualitative-action-transfer} and~\ref{fig:qualitative-rollouts} illustrate reference-action transfer and control across worlds. Transferred latent actions preserve camera or character motion across changes in scene appearance, while keyboard commands drive navigation in real scenes. Additional examples appear in Appendix~\ref{app:additional-qualitative}.

\section{Conclusion}
\label{sec:conclusion}
We presented \onela{}, a shared latent action interface for world models across games, navigation, driving, and robotics. An orthonormal motion dictionary with variational and sparsity regularization structures motion representations learned from unlabeled videos. Lightweight command mappings connect native controls to this space, enabling unified video and action prediction from labeled and unlabeled data. Experiments show strong action readout, cross-view transfer, and action-cycle consistency, while the ablations support using latent actions to learn from unlabeled videos. Together, these results suggest that a common action representation can connect diverse control interfaces without requiring separate world-model backbones. Future work will extend this interface to longer-horizon planning and closed-loop interaction, and test whether it can adapt efficiently to new environments and embodiments.

\bibliography{references}
\bibliographystyle{unsrtnat}

\clearpage
\appendix
\section*{Appendix}

\section{Implementation and Reproducibility Details}
\label{app:implementation}
This appendix details the training configurations and evaluation protocols. Training schedules specify optimization budgets; the checkpoints used for the reported comparisons are identified separately in Appendices~\ref{app:probe-protocol} and~\ref{app:wm-results}.

\subsection{Latent Action Pretraining Video Composition}
\label{app:training-data}
Figure~\ref{fig:onela-data} groups the 1,793,552 latent action pretraining clips into eight video categories. Egocentric pools Ego4D~\citep{grauman2022ego4d}, EgoDex~\citep{hoque2025egodex}, daily activities, and in-the-wild first-person videos; driving combines NVIDIA's PhysicalAI Autonomous Vehicles dataset\footnote{\url{https://huggingface.co/datasets/nvidia/PhysicalAI-Autonomous-Vehicles}} (NVIDIA driving) and nuScenes. Object-centric groups Something-Something v2 (SSv2)~\citep{goyal2017something} interactions and WildRGB-D~\citep{xia2024wildrgbd} object captures, while scene exploration groups Sekai, Spatial-Vid~\citep{wang2025spatialvid}, RealEstate10K, and DL3DV~\citep{ling2024dl3dv}. Robot manipulation combines RT-1 and the two AgiBot collections; handheld UMI~\citep{chi2024umi} contains FastUMI~\citep{zhaxizhuoma2024fastumi}.

Table~\ref{tab:training-data} gives the source-level composition. Game videos come from ABot-World~\citep{jiang2026abotworld0}; the Daily activities, In the wild, and Cinematic scenes sources are collected in-house. We retain 20\% of Spatial-Vid, 50\% of cinematic scenes, and 5\% of daily activities, and use all records from the other configured training manifests. For a manifest with $N$ records and retention ratio $r$, the retained count is $\max(1,\lfloor rN\rfloor)$. Counts refer to video-clip records after this subsampling. Shares use the total retained count as the denominator; they do not denote frame-pair sampling probabilities.

\begin{table}[!htbp]
\centering
\caption{\textbf{Latent action pretraining video composition.} Retained clips and their shares after source-specific subsampling.}
\label{tab:training-data}
\small
\setlength{\tabcolsep}{3pt}
\begin{tabular*}{\linewidth}{@{\extracolsep{\fill}}llrrr@{}}
\toprule
Category & Source & Retained (\%) & Clips & Share (\%) \\
\midrule
Egocentric & Ego4D & 100 & 242,712 & 13.53 \\
 & EgoDex & 100 & 7,136 & 0.40 \\
 & Daily activities & 5 & 159,833 & 8.91 \\
 & In the wild & 100 & 15,203 & 0.85 \\
\addlinespace[2pt]
Games & ABot-World & 100 & 333,201 & 18.58 \\
\addlinespace[2pt]
Driving & NVIDIA driving & 100 & 306,912 & 17.11 \\
 & nuScenes & 100 & 700 & 0.04 \\
\addlinespace[2pt]
Object-centric & SSv2 & 100 & 168,913 & 9.42 \\
 & WildRGB-D & 100 & 85,471 & 4.77 \\
\addlinespace[2pt]
Scene exploration & Sekai & 100 & 18,900 & 1.05 \\
 & Spatial-Vid & 20 & 91,029 & 5.08 \\
 & RealEstate10K & 100 & 6,075 & 0.34 \\
 & DL3DV & 100 & 9,376 & 0.52 \\
\addlinespace[2pt]
Cinematic scenes & Cinematic scenes & 50 & 85,123 & 4.75 \\
\addlinespace[2pt]
Robot manipulation & RT-1 & 100 & 86,912 & 4.85 \\
 & AgiBot 2026 & 100 & 20,324 & 1.13 \\
 & AgiBot & 100 & 90,757 & 5.06 \\
\addlinespace[2pt]
Handheld UMI & FastUMI & 100 & 64,975 & 3.62 \\
\midrule
\multicolumn{3}{l}{Total} & 1,793,552 & 100.00 \\
\bottomrule
\end{tabular*}
\end{table}

\subsection{Latent Action Model}
\label{app:lam-training}
The 32-dimensional LAM uses transformer width 1024, patch size 16, 16 attention heads, and 24 blocks in each encoder and decoder stack. Spatial rotary position embeddings and scaled dot-product attention are enabled. The loss uses KL coefficient $\beta=10^{-6}$ and sparsity coefficient $\lambda=0.01$, with no auxiliary contrastive loss. Training samples from the diagonal Gaussian posterior; representation extraction uses its mean. The decoder adds the embedded action to each current-frame patch.

\paragraph{Optimization.}
We train from scratch on 16 GPUs using fully sharded data parallelism and AdamW with a learning rate of $5\times10^{-5}$, weight decay 0.01, and gradient clipping at 0.3. Each GPU processes 40 frame pairs, giving a global batch of 640 without gradient accumulation. The cosine schedule decays to $5\times10^{-6}$. The default schedule specifies five passes over a 6.4M-pair sampling budget, for up to 50,000 updates. The data-amount and composition comparisons below instead use constant learning rates. Mixed precision is disabled and the random seed is 42.

\paragraph{Frame-pair sampling.}
Frames are resized to height 240 with aspect ratio preserved, patch-aligned widths, and a maximum width of 864. Most sources are sampled at 16 frames per second (FPS) with frame offsets in $\{0,1,2,4,8,16\}$. RT-1 uses 3 FPS and offsets $\{0,1,2,3,6\}$; SSv2 and nuScenes use 12 FPS, and NVIDIA driving and daily-activity videos use 10 FPS. nuScenes additionally includes offset 12. Pair starting positions advance by two sampled frames except for RT-1, which uses stride one. Zero-offset pairs provide static transitions. Per-video pair limits range from 10 to 40 across sources. Source selection uses the configured sample-count sampler after the manifest subsampling in Appendix~\ref{app:training-data}.

\paragraph{Model-size and data-amount comparisons.}
\label{app:model-scaling}
Layers in Table~\ref{tab:model-data-scaling}(a) count transformer blocks in each stack: 6/6, 12/12, and 24/24 encoder/decoder blocks. All use width 1024, 16 attention heads, 16-pixel patches, and 32 latent coefficients. Parameter counts include the encoder, decoder, projections, action prompt, and motion dictionary: 180,870,400, 357,172,672, and 709,777,216, respectively. They exclude the downstream video world model and native-command MLPs. Table~\ref{tab:model-data-scaling}(b) uses three checkpoints from the same 24/24-layer training run with a constant learning rate of $5\times10^{-5}$ and a global batch of 640 frame pairs. The $1\times$, $2\times$, and $3\times$ settings correspond to 10,000, 20,000, and 30,000 updates, or 6.4M, 12.8M, and 19.2M sampled frame pairs. The source mixture is fixed, and data amount measures cumulative sampled frame pairs.

\paragraph{Data-composition comparisons.}
\label{app:data-composition}
The four mixtures in Table~\ref{tab:ablation-data-types} use 24/24-layer LAMs with 32-dimensional coefficients and a constant learning rate of $5\times10^{-5}$. The evaluation uses the same six-dataset linear-probe protocol.

\subsection{World-Model Training}
\label{app:wm-training}
\paragraph{Training data and native-label usage.}
The world-model mixture comprises the 11 sources in Table~\ref{tab:wm-training-data}, distinct from the larger LAM pretraining collection. With a global batch of 16 clips over 6,000 updates, world-model training samples 96,000 clips. Native action annotations and matching command MLPs are available for ABot-World~\citep{jiang2026abotworld0}, AgiBot, nuScenes, and RT-1; the remaining sources use frame-pair-derived latent actions.

\begin{table}[!htbp]
\centering
\caption{\textbf{World-model training data and native-label usage.} Expected sample counts over 6,000 updates with batch size 16, rounded to integers. Labeled counts are before action dropout.}
\label{tab:wm-training-data}
\small
\setlength{\tabcolsep}{4pt}
\begin{tabular*}{\linewidth}{@{\extracolsep{\fill}}lrr@{}}
\toprule
Source & Sample count & Labeled count \\
\midrule
ABot-World & 42,422 & 8,484 \\
AgiBot & 16,204 & 3,241 \\
nuScenes & 1,250 & 250 \\
RT-1 & 4,490 & 898 \\
\addlinespace[2pt]
DL3DV & 1,674 & 0 \\
EgoDex & 2,548 & 0 \\
NVIDIA driving & 5,447 & 0 \\
RealEstate10K & 7,866 & 0 \\
Sekai & 3,375 & 0 \\
Spatial-Vid & 7,708 & 0 \\
SSv2 & 3,016 & 0 \\
\midrule
Total & 96,000 & 12,873 \\
\bottomrule
\end{tabular*}
\end{table}

Counts are proportional to each source's share of the training mixture and include repeated sampling. Labeled count denotes the expected number of samples using native controls during world-model fine-tuning, totaling approximately 12,873 (13.41\%). Latent-action prediction targets are extracted by the frozen LAM for both conditioning routes.

\paragraph{Initialization and conditioning.}
The video model uses a pretrained Wan2.2-TI2V-5B~\citep{wan2025modelcard} checkpoint from ABot-World~\citep{jiang2026abotworld0} and shares its backbone weights across domains. The LAM is frozen, and four pretrained command MLPs and an additive action adapter initialize the conditioning path. Each MLP has two linear layers with a sigmoid linear unit (SiLU) nonlinearity, hidden width 256, and no input layer normalization. Alignment targets are posterior means from the frozen LAM. Four consecutive 32-dimensional actions are concatenated into each 128-dimensional adapter input (Pack1; see below). Clips with native controls use either mapped controls or frame-pair-derived latent actions; clips without native controls use frame-pair-derived latent actions. The MLPs remain trainable. The joint objective combines video flow matching with latent action prediction, with action-loss weight 0.05 and cosine-loss weight 0.1.

\paragraph{Native-control alignment.}
Before world-model training, each MLP regresses frozen-LAM posterior means using SmoothL1 loss plus cosine distance weighted by 0.1. The keyboard, driving, AgiBot, and RT-1 inputs have 8, 6, 14, and 7 dimensions, respectively, and all outputs have 32 dimensions. Pre-alignment uses AdamW with a constant learning rate of $10^{-4}$, weight decay 0.01, and a global batch of 32 on four GPUs, with a budget of 20,000 updates. AgiBot's left and right gripper deltas are divided by training-set scales 30.6821 and 30.5333; other inputs retain their native scale. These MLPs initialize the subsequent joint training and remain trainable.

\paragraph{Auxiliary action prediction.}
The action head reads spatial features from adjacent DiT temporal token groups. Learned queries cross-attend to these features, and an output projection regresses four consecutive 32-dimensional actions per transition between video-latent steps. Targets are detached posterior means from the frozen LAM. The loss combines SmoothL1 regression with cosine distance computed separately for each 32-dimensional action, weighted by 0.1; the combined action loss is weighted by $\lambda_{\mathrm{act}}=0.05$. For video-derived conditioning, the same inferred codes provide both the condition and the auxiliary supervision target.

\paragraph{Optimization and video sampling.}
World-model training uses 16 GPUs and AdamW with a global batch of 16 clips, a learning rate of $10^{-5}$, AdamW coefficients $(0.9,0.99)$, and weight decay 0.01. We fine-tune for 6,000 updates with cosine learning-rate decay (Appendix~\ref{app:wm-results}). Gradients are clipped at 1.0; bfloat16 mixed precision and gradient checkpointing are enabled, and the seed is 42. Clips contain 121 frames. The video reader resizes the short side to 704 pixels, preserves the aspect ratio, and aligns spatial dimensions to multiples of 32. The target frame rate is 16 FPS, with MIND and RT-1 retaining their native frame rates. LAM conditioning images use the separate height-240 preprocessing above and are encoded in microbatches of eight. Action conditioning is dropped with probability 0.1. Reference-image conditioning uses up to five slots at resolution 512 and a drop probability of 0.1. Clip-level and frame-level static augmentation probabilities are 0.03 and 0.02; horizontal flipping is disabled.

\subsection{Video and Control Alignment}
For $T+1$ sampled frames, consecutive pairs provide $L=T$ action conditions and targets. In the reported \emph{Pack1} setting, the LAM encodes consecutive frame pairs; each stride-four video-latent step receives four concatenated 32-dimensional vectors. The resulting 128-dimensional condition preserves the ordering of the four transitions. This is different from encoding only the endpoints four frames apart. Native control $u_i$ describes the transition arriving at sampled frame $i$: the transition from frame 0 to frame 4 uses controls $u_1,\ldots,u_4$. Mapped controls follow the same packing. The initial conditioning frame has no newly generated transition and is excluded from fidelity scores.

\section{Probe Labels and Statistical Reporting}
\label{app:labels}
Table~\ref{tab:domains} summarizes the sample counts and action labels used in the frozen-representation evaluation.

\begin{table}[htbp]
\centering
\caption{\textbf{Frozen-representation evaluation sets.} Counts refer to frame pairs. Binary labels are predicted independently. Sekai macro-F1 averages the six labels with positive test examples; I/K have no positive examples.}
\label{tab:domains}
\small
\setlength{\tabcolsep}{4pt}
\begin{tabular*}{\linewidth}{@{\extracolsep{\fill}}lrrl@{}}
\toprule
Dataset & Fit & Test & Action labels \\
\midrule
Game & 800 & 1,600 & 8 binary keyboard controls \\
Sekai & 607 & 1,222 & 6 evaluated motion labels \\
MIND & 1,500 & 1,500 & 8 actions; first-/third-person views \\
nuScenes & 2,342 & 2,441 & 4 binary motion directions \\
RT-1 & 1,640 & 1,650 & 14 signed motion/gripper labels \\
AgiBot & 1,600 & 1,600 & 28 signed labels across two arms \\
\bottomrule
\end{tabular*}
\end{table}

\subsection{Reported Frozen-Representation Protocol}
\label{app:probe-protocol}
\paragraph{Checkpoints and preprocessing.}
We compare the 32-dimensional \onela{} with OLaF~\citep{jiang2026olaf}, its VAE checkpoint without alignment, DiLA~\citep{zhang2026dila}, and the ManiSkill checkpoint of RLA-WM~\citep{zhang2026rla}. AdaWorld~\citep{gao2025adaworld} uses its released 32-dimensional LAM and posterior-mean representations. All methods use the same evaluation pairs and dataset-specific readout protocol, retaining their native representation dimensions and input preprocessing. DiLA uses a 256-dimensional code; RLA-WM uses flattened residual features with 2,048 dimensions.

\paragraph{Readouts and evaluation splits.}
For \onela{}, we use a consistent video-level train/test partition across LAM pretraining and world-model training. All test videos used in representation and world-model evaluations are excluded from both training stages. Our primary metric is macro-F1 from a single linear layer over the full representation. For MIND, we follow the train/test partition of Olaf-World~\citep{jiang2026olaf}. Our MIND probe is an eight-way classifier; the other datasets use independent binary classifiers. We also test a restricted readout that selects one dimension per action by its absolute correlation with fit labels, then fits a sign and decision threshold. This \emph{single-dimension} test measures how directly individual dimensions expose action information; it still requires labeled fit data. For nuScenes, we report full-representation probing. Within each dataset, all methods use the same probe-fitting protocol. AgiBot uses disjoint sets of 200 fit and 200 test episodes.

\paragraph{Cross-view transfer.}
For each MIND transfer direction, we fit an eight-way linear probe on the source-view training subset and apply it unchanged to the target-view test subset. Feature means and standard deviations are estimated only from the source training subset; target-view labels are used only to compute macro-F1.

\paragraph{Reporting.}
Readouts measure accessibility of the provided action labels. Simultaneous binary controls retain their multilabel form, and robot labels use each dataset's native action convention.

\section{Metric and Qualitative Evaluation Contracts}
\label{app:metrics}
\subsection{Reported World-Model Comparison}
\label{app:wm-results}
\paragraph{Checkpoint selection.}
All methods in Table~\ref{tab:wm} are evaluated after 6,000 fine-tuning updates. For \onela{}, MIND, nuScenes, and RT-1 use a single shared checkpoint with the corresponding WASD/IJKL, pose, or RT-1 MLP. Inputs are the initial frame and keyboard commands on MIND, known ego-pose trajectories on nuScenes, or robot actions on RT-1; future target RGB frames are not used to extract action conditions. Cycle transfer extracts latent actions from the source video; the return pass extracts them from the generated transfer video and starts from the original source initial frame. The baselines are fine-tuned on the common 11-source mixture, including RT-1. Their LAMs remain frozen while the video model and action-conditioning components are jointly trained.

\paragraph{Samples and aggregation.}
For \onela{}, test videos are held out from both LAM pretraining and world-model training, following the shared partition in Appendix~\ref{app:labels}. All four methods use the same fixed test cases, 50 denoising steps, guidance scale 3, and seed 42. MIND contains 100 videos of 97 frames at 24 FPS. nuScenes and RT-1 each contain 40 videos. Cycle transfer contains 40 videos, evenly divided among the four source--target--source domain combinations. Metrics are first averaged within each video and then equally across videos. Scores cover all test cases. Relative improvements are computed from unrounded scores.

\paragraph{Visual metrics.}
PSNR (dB), SSIM~\citep{wang2004ssim}, and LPIPS~\citep{zhang2018lpips} with the VGG16 backbone compare predictions and ground truth at the original video geometry, excluding the conditioning frame. IQ is MUSIQ-SPAQ~\citep{ke2021musiq} rescaled to $[0,100]$ for reporting, evaluated on all frames including the first frame with the existing VBench~\citep{huang2024vbench} long-side-512 preprocessing. IQ measures visual quality rather than action correctness.

\paragraph{Trajectory metrics.}
MIND and cycle RPE use ViPE~\citep{huang2025vipe} default pose estimation, Sim(3) alignment of the predicted trajectory to the reference, and adjacent-frame relative poses ($\Delta=1$). Translation error is expressed in the estimated reference-trajectory scale; rotation error is in degrees. MIND uses 24 FPS. Cycle evaluation uses 16 FPS and the source video's actual length (21, 41, or 121 frames), comparing the returned video to the source. RPE is aggregated separately within each benchmark.

\paragraph{Training-configuration and collision comparisons.}
Figure~\ref{fig:wm-training-ablation}(a--c) compares training w/ and w/o latent action after 6,000 fine-tuning updates. Both settings use the same data and training procedure, train the MLPs, and retain the initial frame and native-command pathway. The only difference is whether unlabeled videos receive LAM-inferred latent actions or zero action vectors. The real-scene test uses game-action commands with 30 initial frames, ten each from Sekai, Spatial-Vid, and RealEstate10K. For HPSv3~\citep{ma2025hpsv3}, we uniformly sample 16 frames per generated video, excluding the initial frame and rounding indices to the nearest frame. Each case uses the same fixed target-scene text caption across setups and frames. We average raw preference means ($\mu$) across frames and then equally across videos. This measures visual preference rather than control correctness. The collision comparison in panel (d) is a separate 50-case test of each loss setting. Collision outcomes are verified by human inspection of the generated videos. Collision rates are 36\% (18/50) without the latent-action prediction loss and 22\% (11/50) with it.

\paragraph{Qualitative example provenance.}
Each control example shows the initial frame followed by five generated frames, with labels from its command record at the displayed frame indices. A key labels the action arriving at that frame, not a constant action over the interval between displayed frames; the initial frame has no generated transition. Each transfer example uses the same three frame indices for the source and transfer videos; the first transfer frame is the generation input.

\section{Orthogonal Parameterization: Scope of the Constraint}
\label{app:geometry}
Equation~\eqref{eq:isometry} follows directly from $Q^\top Q=I$: for $v=z-z'$, $\|Qv\|_2^2=v^\top Q^\top Qv=\|v\|_2^2$. This is a property of the embedding map, not of the generated transition. A nonlinear decoder can respond differently to the same coefficient change at different observations. Moreover, for an orthogonal $U\in\R^{d\times d}$, the pair $(QU,U^\top z)$ produces the same embedded action. The isotropic Gaussian prior does not resolve that rotational ambiguity. Sparsity changes the preference among coordinate systems but does not identify a unique physical basis.

\clearpage
\section{Additional Qualitative Results}
\label{app:additional-qualitative}
\begin{figure}[!ht]
\centering
\includegraphics[page=1,width=\linewidth]{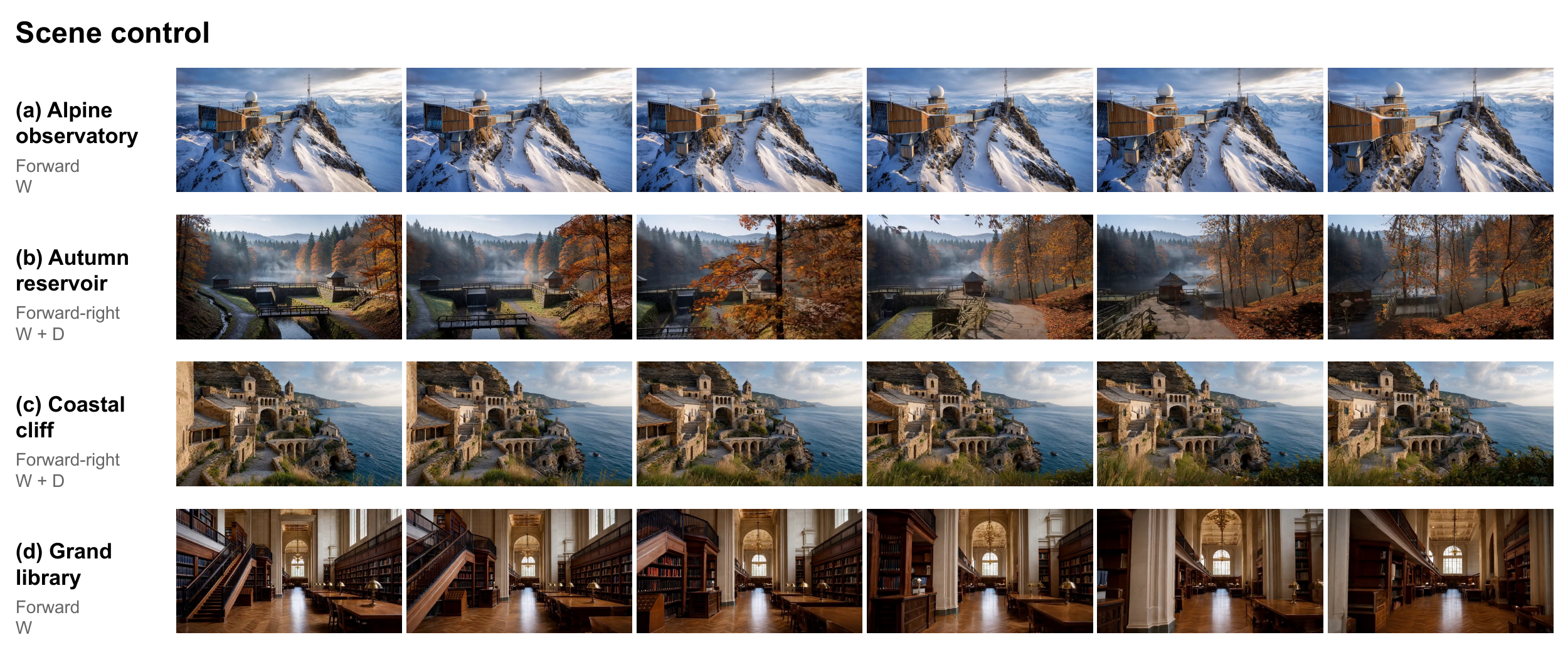}
\caption{\textbf{Additional \onela{} scene-control rollouts.} Six frames per case illustrate motion under the indicated keyboard controls.}
\label{fig:supp-scene-control-a}
\end{figure}

\begin{figure}[!ht]
\centering
\includegraphics[page=2,width=\linewidth]{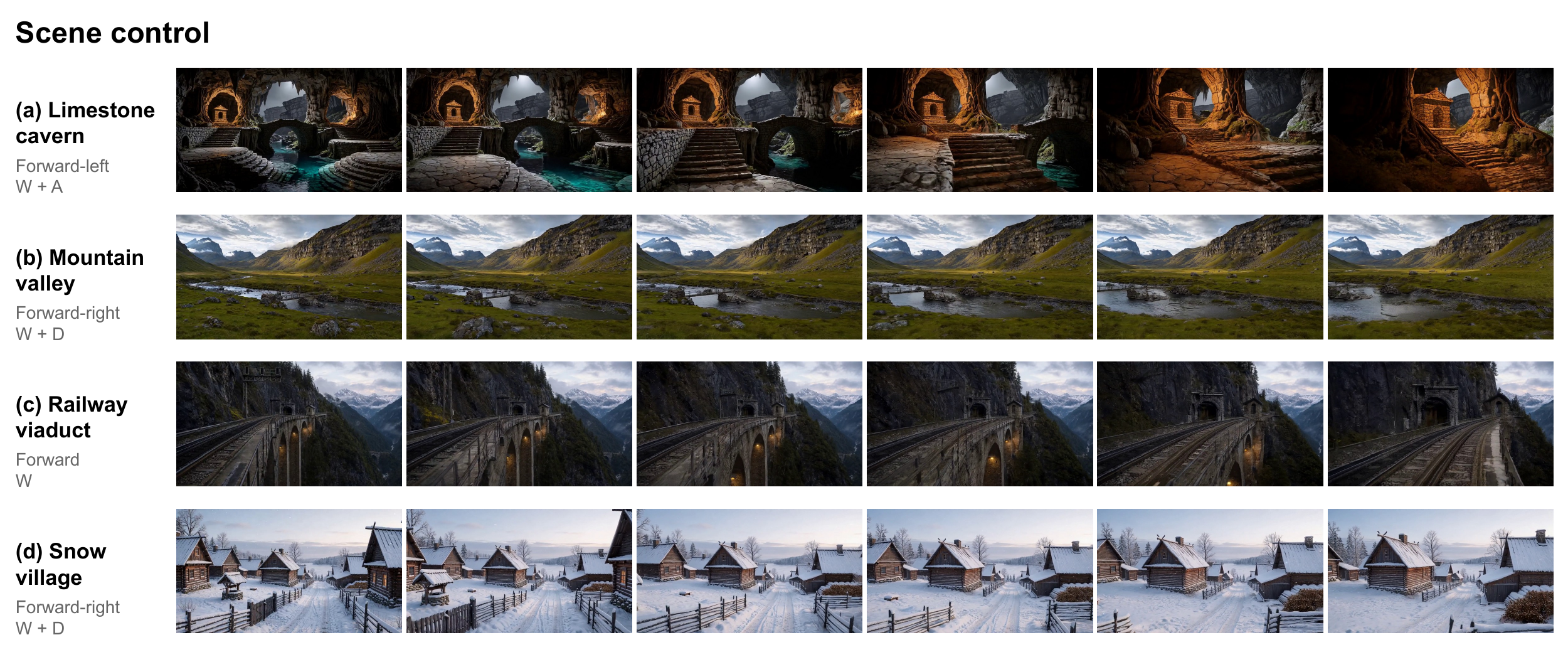}
\caption{\textbf{Additional \onela{} scene-control rollouts in natural and built environments.} Each row follows the indicated keyboard control.}
\label{fig:supp-scene-control-b}
\end{figure}

\begin{figure}[!ht]
\centering
\includegraphics[page=1,width=\linewidth]{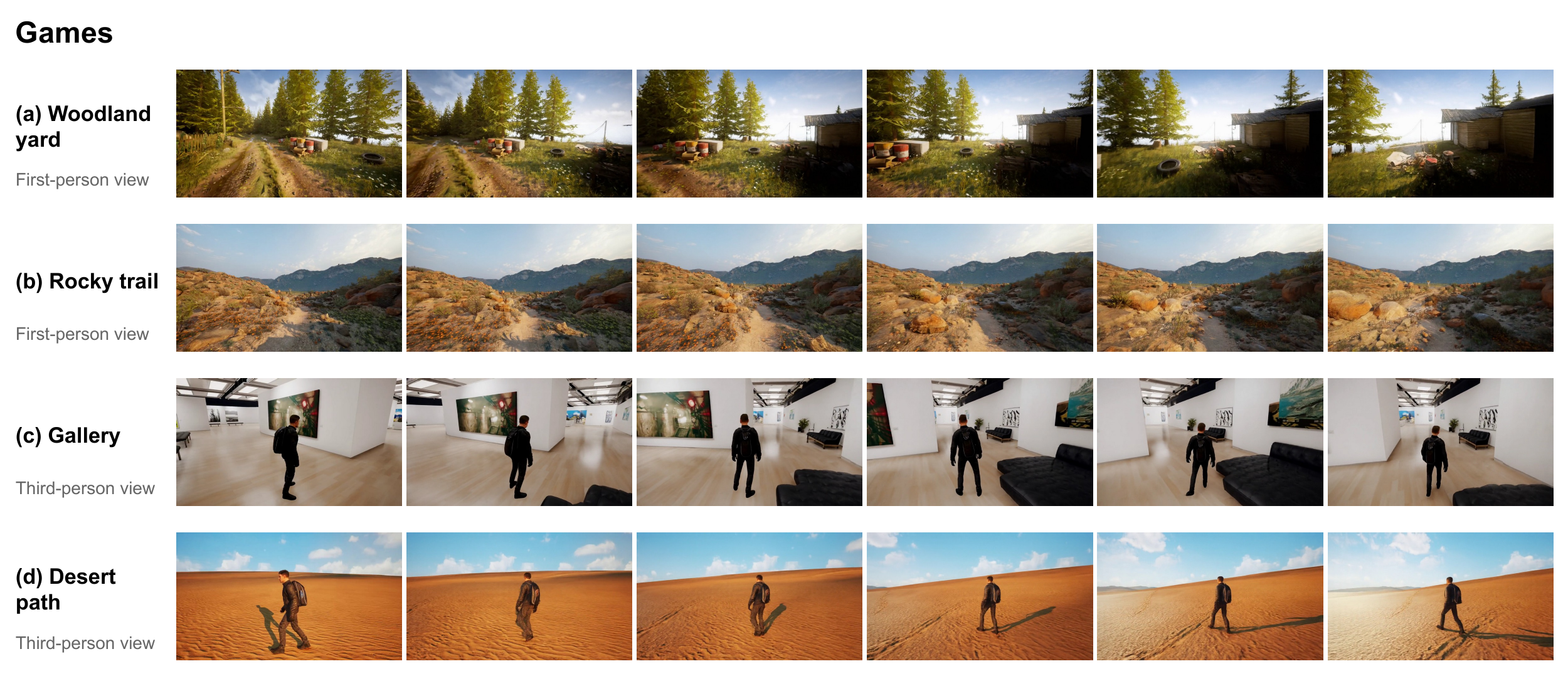}
\caption{\textbf{Additional \onela{} rollouts in games.} Examples include first- and third-person views, with six frames sampled from each rollout.}
\label{fig:supp-games}
\end{figure}

\begin{figure}[!ht]
\centering
\includegraphics[page=2,width=\linewidth]{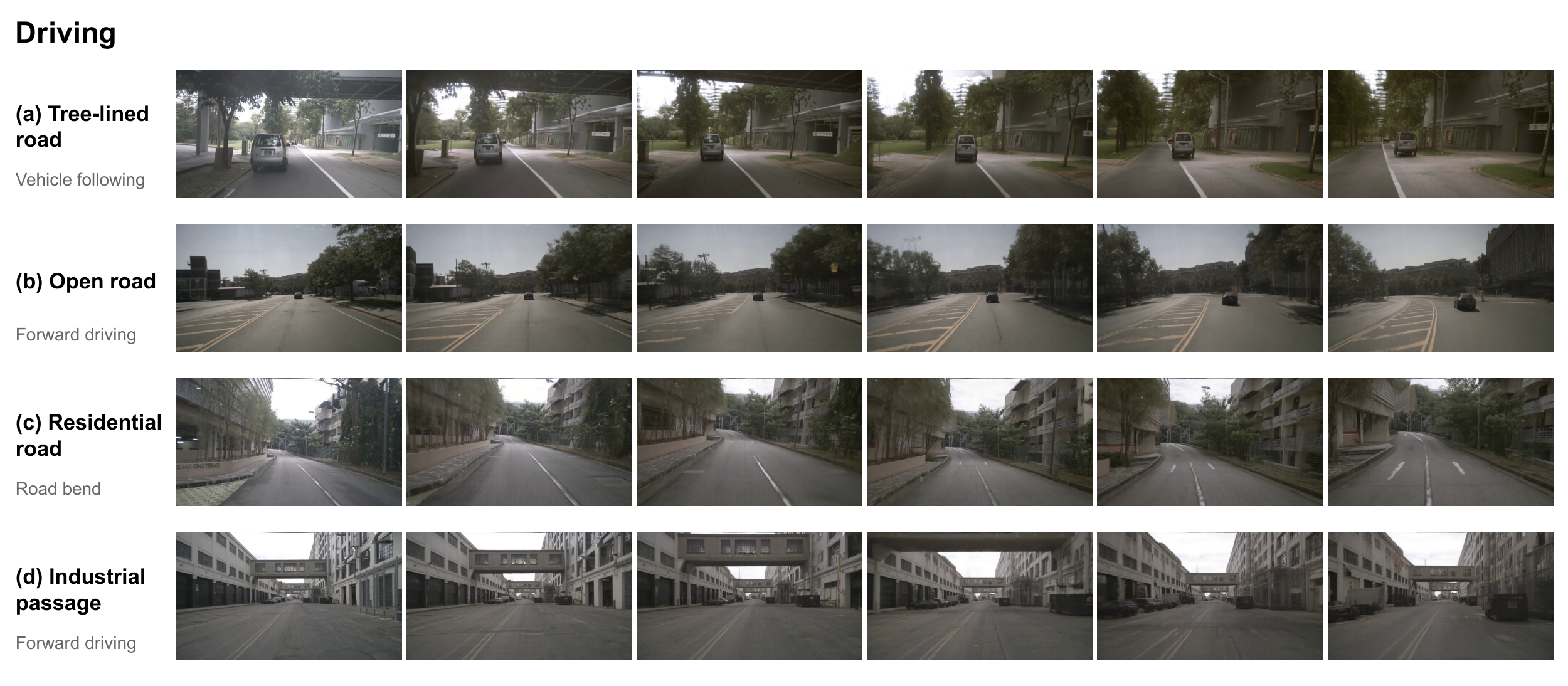}
\caption{\textbf{Additional \onela{} driving rollouts.} Examples cover vehicle following, open roads, a road bend, and an industrial passage.}
\label{fig:supp-driving}
\end{figure}

\begin{figure}[!ht]
\centering
\includegraphics[page=3,width=\linewidth]{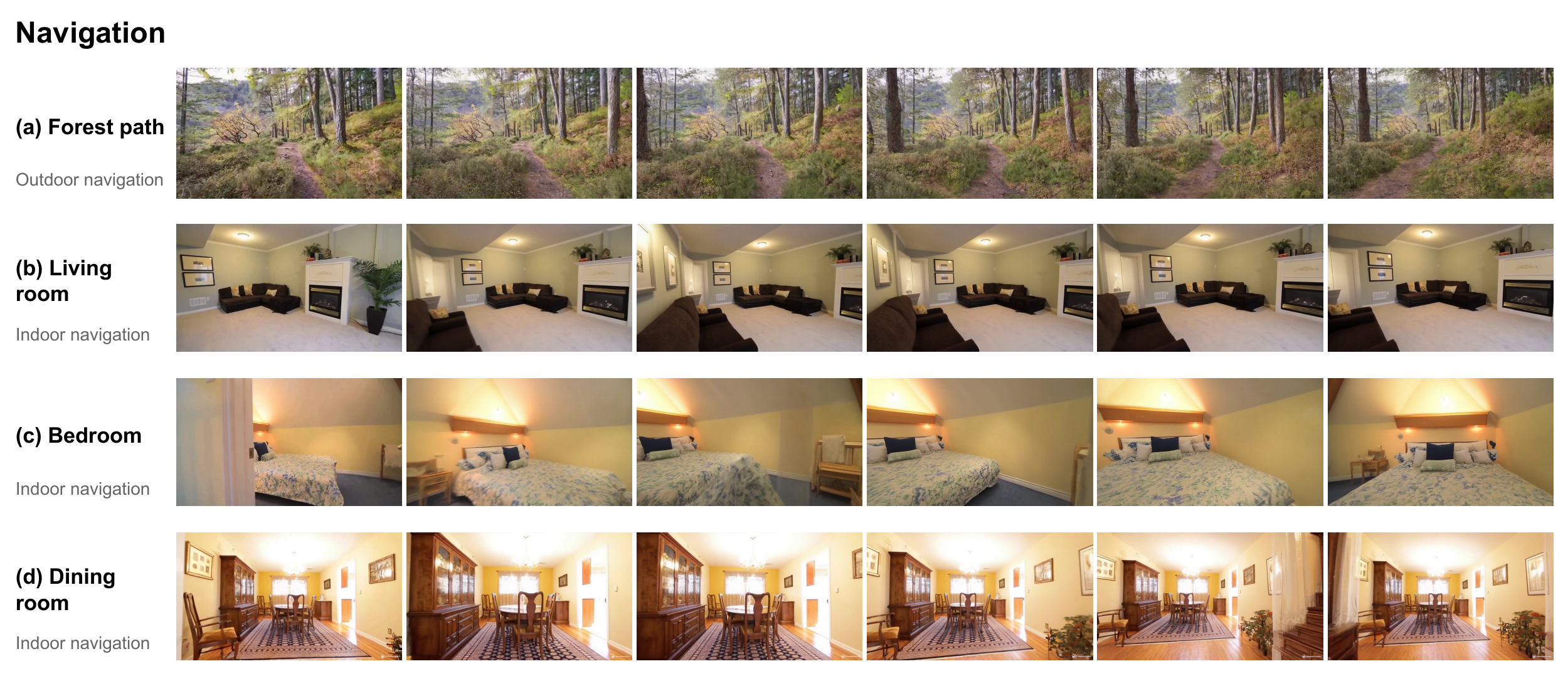}
\caption{\textbf{Additional \onela{} navigation rollouts.} Examples show camera motion through outdoor and indoor scenes.}
\label{fig:supp-navigation}
\end{figure}

\begin{figure}[!ht]
\centering
\includegraphics[page=4,width=\linewidth]{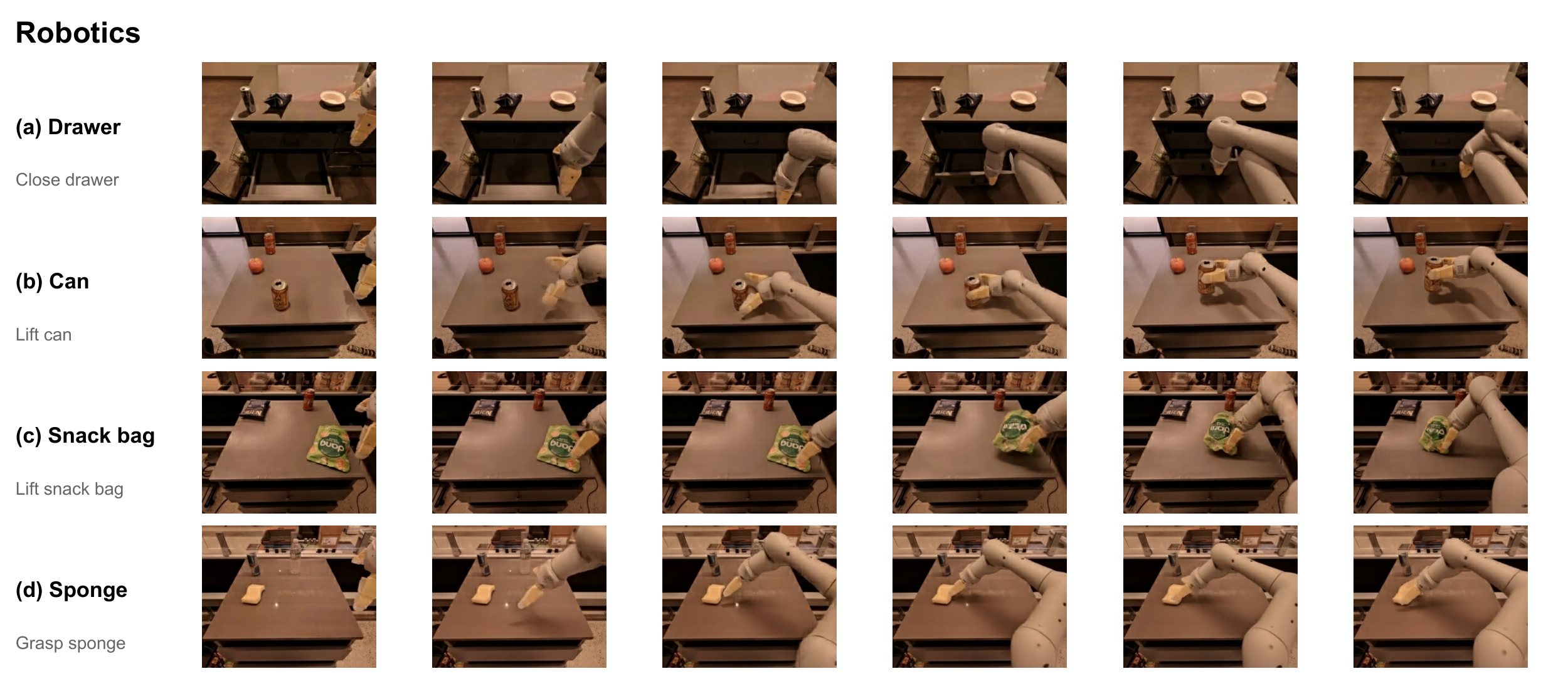}
\caption{\textbf{Additional \onela{} robot rollouts.} Examples show interactions with a drawer, a can, a snack bag, and a sponge.}
\label{fig:supp-robotics}
\end{figure}

\begin{figure}[!ht]
\centering
\includegraphics[page=1,width=\linewidth]{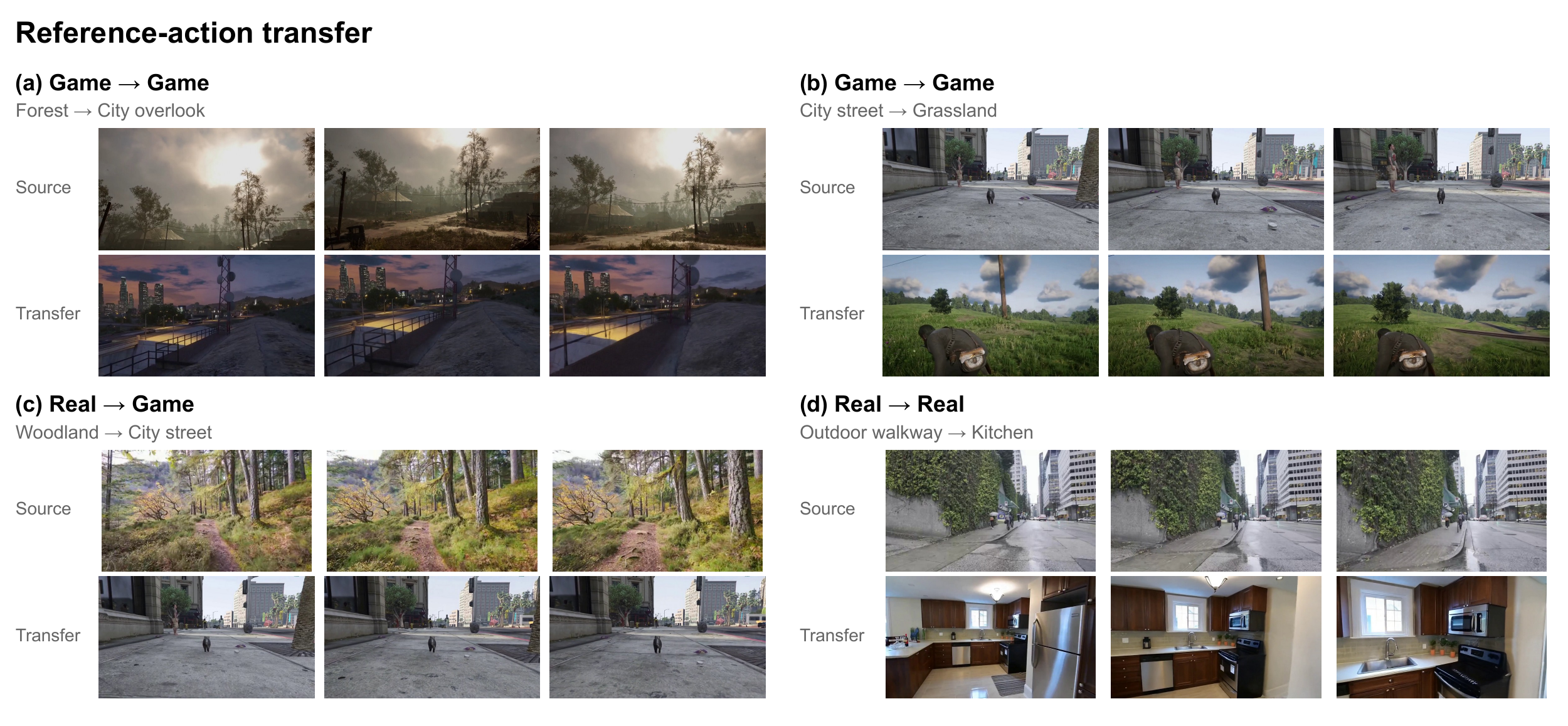}
\caption{\textbf{Additional \onela{} reference-action transfer results.} Each case pairs a source video (top) with a generated rollout in a different scene (bottom). Columns show matched temporal positions.}
\label{fig:supp-action-transfer}
\end{figure}

Figures~\ref{fig:supp-scene-control-a} and~\ref{fig:supp-scene-control-b} show additional scene-control rollouts. Further examples cover games (Figure~\ref{fig:supp-games}), driving (Figure~\ref{fig:supp-driving}), navigation (Figure~\ref{fig:supp-navigation}), and robotics (Figure~\ref{fig:supp-robotics}); Figure~\ref{fig:supp-action-transfer} shows reference-action transfer. Frames are ordered from left to right; source and transfer frames are temporally paired.

\subsection{Scene Control}

\clearpage
\subsection{Rollouts across Domains}

\clearpage

\clearpage
\subsection{Reference-Action Transfer}

\clearpage
\section{Enlarged Cross-Domain Similarity Matrices}
\label{app:enlarged-qr}
\begin{figure}[!ht]
\centering
\includegraphics[width=0.95\linewidth]{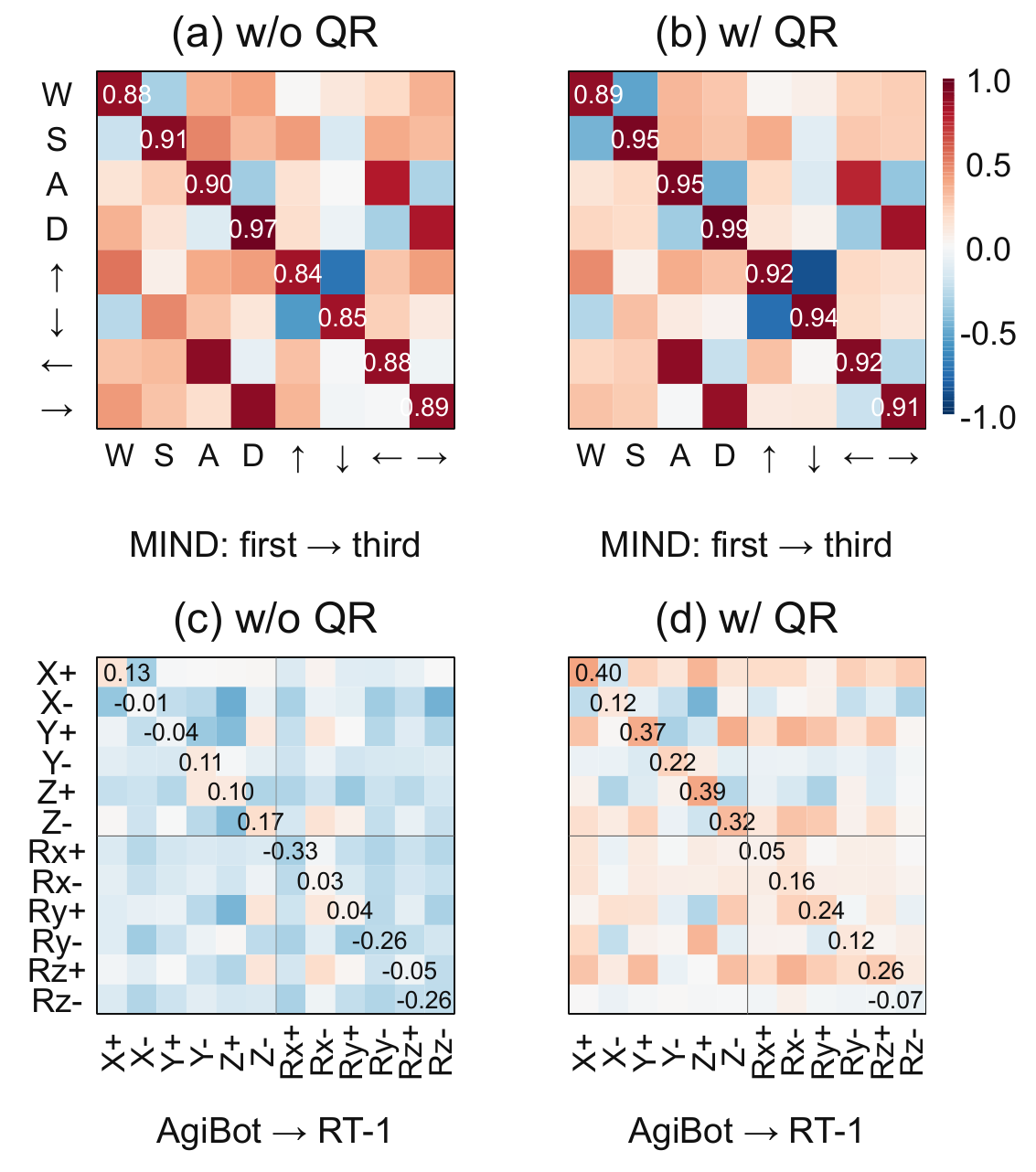}
\caption{\textbf{Enlarged view of Figure~\ref{fig:qr-cross-domain}.} Diagonal annotations give the cosine similarity between prototypes with matching action labels.}
\label{fig:qr-cross-domain-enlarged}
\end{figure}

Figure~\ref{fig:qr-cross-domain-enlarged} provides an enlarged view of the cross-domain similarity matrices in Figure~\ref{fig:qr-cross-domain}.

\end{document}